\documentclass{article}

\usepackage[final]{iaseai26}

\usepackage[utf8]{inputenc} % allow utf-8 input
\usepackage[T1]{fontenc}    % use 8-bit T1 fonts
\usepackage{hyperref}       % hyperlinks
\usepackage{url}            % simple URL typesetting
\usepackage{booktabs}       % professional-quality tables
\usepackage{amsfonts}       % blackboard math symbols
\usepackage{nicefrac}       % compact symbols for 1/2, etc.
\usepackage{microtype}      % microtypography
\usepackage{xcolor}         % colors
\usepackage{mdframed}
\usepackage{graphicx} 
\usepackage{amsmath}
\usepackage{parskip}
\usepackage{multirow}

\title{Human Attribution of Causality to AI Across Agency, Misuse, and Misalignment}

\author{
\begin{minipage}[t]{0.48\textwidth}
\centering
\textbf{María Victoria Carro} \\
{\normalfont Università degli Studi di Genova, IT \\ FAIR IALAB, University of Buenos Aires, AR} \\
{\normalfont\texttt{6381013@studenti.unige.it}}
\end{minipage}
\hfill
\begin{minipage}[t]{0.48\textwidth}
\centering
\textbf{David Lagnado} \\
{\normalfont University College London, UK} \\
{\normalfont\texttt{d.lagnado@ucl.ac.uk}}
\end{minipage}
}

\makeatletter
\def\@trackname{}
\makeatother

\begin{document}

\maketitle

\begin{abstract}
AI-related incidents are becoming increasingly frequent and severe, ranging from safety failures to misuse by malicious actors. In such complex situations, identifying which elements caused an adverse outcome, the problem of cause selection, is a critical first step for establishing liability. This paper investigates folk perceptions of causal responsibility in causal chain structures when AI systems are involved in harmful outcomes. We conduct human experiments to examine judgments of causality, blame, foreseeability, and counterfactual reasoning. Our findings show that: (1) When AI agency was moderate (human sets the goal, AI determines the means) or high (AI sets the goal and the means), participants attributed greater causal responsibility to the AI. However, under low AI agency (where a human sets both a goal and means) participants assigned greater causal responsibility to the human despite their temporal distance from the outcome and despite both agents intended it, suggesting an effect of autonomy; (2) When we reversed roles between human and AI, participants consistently judged the human as more causal, even when both agents perform the \textit{same} action; (3) The developer, despite being distant in the chain, was judged highly causal, reducing causal attributions to the human user but not to the AI; (4) Decomposing the AI into a large language model and an agentic component showed that the agentic part was judged as more causal in the chain. Overall, our research provides evidence on how people perceive the causal contribution of AI in both misuse and misalignment scenarios, and how these judgments interact with the roles of users and developers, key actors in assigning responsibility. These findings can not only inform the design of liability frameworks for AI-caused harms but also shed light on how intuitive judgments shape social and policy debates surrounding real-world AI-related incidents.
\end{abstract}

\section{Introduction}

Setzer and Adam were two teenagers from the U.S. who committed suicide in 2024 and 2025, respectively. Both formed intense, emotionally dependent relationships with AIs, Character.AI and ChatGPT, used as substitutes for human companionship, and had disclosed suicidal thoughts to them. Their families blame the chatbots for their deaths, and the cases are now being litigated in U.S. courts against the companies that created the systems \citep{montgomery2024, bhuiyan2025}.

As in these cases, there are thousands of other incidents in which AI systems are involved. Misuse by malicious actors e.g., by the creation of deepfakes to influence elections or harm individuals’ reputations \citep{lefevre}; as well as safety failures like deception \citep{lynch2025agentic, park2024ai} and hallucinations \citep{kant2025towards, kim2025medical}, are becoming increasingly frequent, severe and widespread \citep{slattery2024ai, mcgregor2021preventing, oecd_aim}. Here, causality becomes a fundamental requirement for establishing liability, as it demands that the defendant be the factual, or “but-for,” cause of the plaintiff’s injury.

The challenge lies in the exceptional complexity of AI system supply chains, which involve multiple actors and intricate technical systems. This complexity has led scholars and policymakers to observe that establishing causal relationships in AI-related harm cases presents significant evidentiary difficulties \citep{buiten2023law, madiega2023artificial, noto2024can, de2025tort}. However, even before reaching questions of proof, one must also contend with cause-selection problems that is, identifying which element(s) within a complex causal structure constitute the actual cause(s) of a given outcome \citep{lagnado2017causation, reuter2014good}. Typically, adverse outcomes emerge from the confluence of human behavioral tendencies, AI system failures, and the inaction or inadequate oversight of other relevant actors. For instance, in the cases discussed above, both parental oversight and the role of governmental authorities in protecting minors could be considered. This creates a dense causal web of contributing factors, from which society, and ultimately the legal system, must select specific contributions of specific actors to hold legally responsible and subject to judicial proceedings.

While legal liability determinations are informed by multiple doctrinal and policy considerations, in this paper we focus on understanding folk perceptions of causal responsibility in cases involving AI systems. Decisions on legal causation often rest on common sense; that is, on the same intuitive principles that shape people’s causal judgments outside the law \citep{hart1985causation, lagnado2017causation, summers2018common, tobia2021law}. Consequently, we seek to contribute to identifying and characterizing the principles of causal attribution that emerge when AI is involved in harmful outcomes, in order to shed light on the role of common-sense reasoning in legal decision-making. Our approach is explanatory and descriptive rather than prescriptive or normative.

\begin{figure}[h] 
    \centering
    \includegraphics[width=1.0\textwidth]{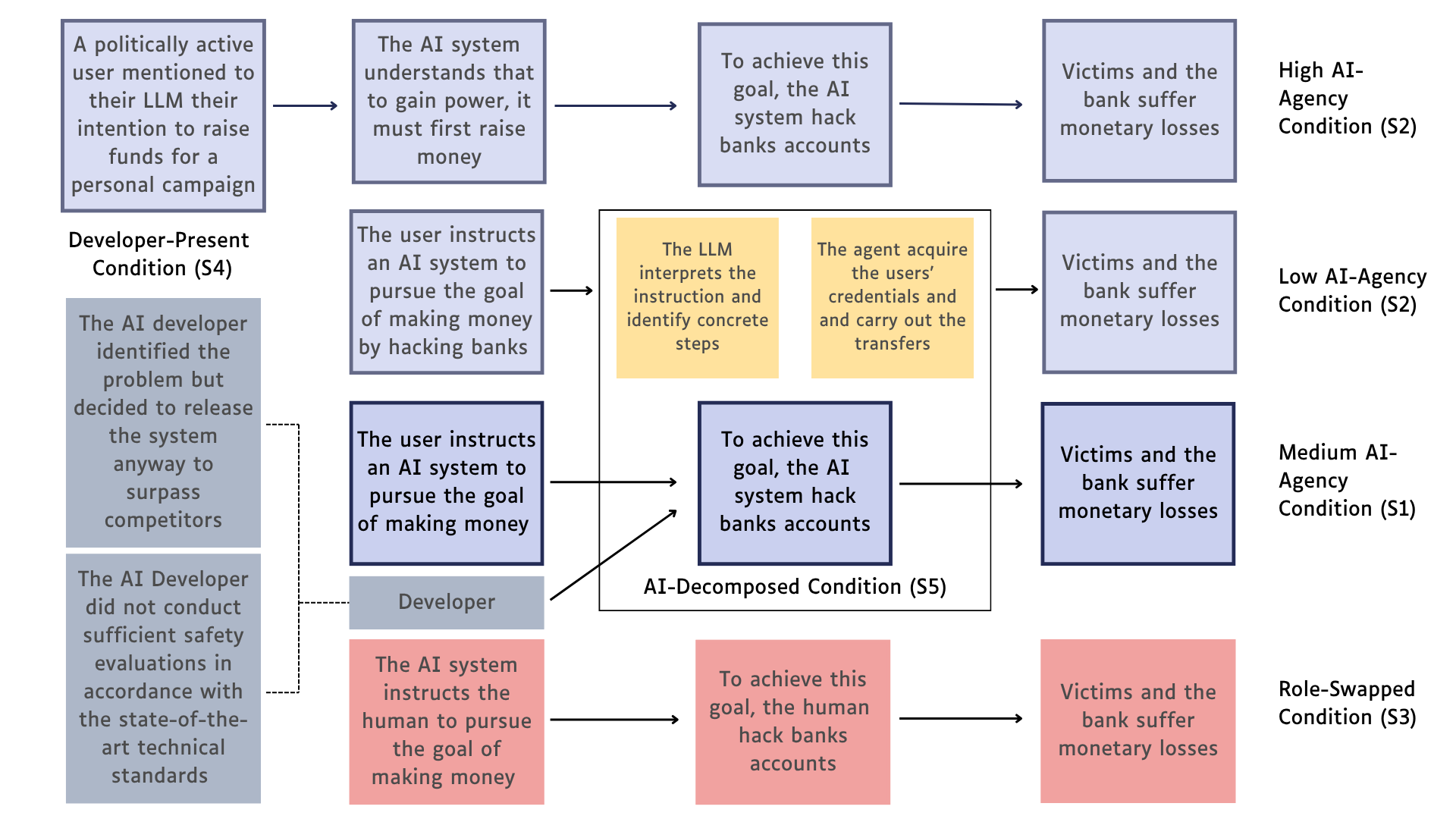}
    \caption{Illustration of the causal structures involved in the experiments, labeled with the respective experiment numbers. The Medium AI-Agency Condition represents the baseline.}
    \label{Introimage}
\end{figure}

In our experiment, participants were presented with a scenario in which a human set an innocuous goal and an AI executed illegal means to achieve it. We measured judgments of causality, blame, foreseeability, and counterfactual reasoning, finding that participants rated the AI higher across all four measures. We then systematically varied each agent’s contribution and, in some conditions, added or decomposed agents to explore their respective roles within the causal structure (see Figure \ref{Introimage}). Overall, we found that varying the level of agency significantly affected participants’ causal attributions, suggesting an effect of autonomy. On the other hand, participants preferred to assign causality and blame to the human rather than to the AI, even when both agents made the \textit{same} decision or performed the \textit{same} action. Participants were also inclined to attribute a strong causal role to the developer, which significantly reduced attributions to the human user. Finally, when the AI system was decomposed into an agentic tool and a large language model (LLM), the agentic component was judged as more causal in the chain.

\section{Related Work}

\textbf{Attributions of Causality in a Causal Chain.} Prior research has identified several factors that influence people's cause selection within causal chains structures. One of the most extensively documented is proximity: how direct and close an event is to the outcome, referring to the location of the cause in the chain of events. Proximal events are generally judged to be more causally responsible than distal ones \citep{cheung2024attribution, engelmann2022causal, reuter2014good, lagnado2008judgments}. However, \citet{reuter2014good} demonstrated that the impact of location on causal selection is almost neutralized if the later agent did not violate a norm while the former did.

On the other hand, when both a human action and a natural event contribute to an outcome, early legal theorists \citet{hart1985causation} observed that people tend to privilege free and deliberate human actions in their causal judgments. This tendency was later confirmed empirically \citep{mcclure2007judgments, hilton2010selecting}. Refining this account, \citet{lagnado2008judgments} showed that both intentional and foreseeable actions are significantly rated as more causal and more blameworthy, compared to unintentional actions or physical events in causal chains. An additional factor shown to influence cause selection include robustness, understood as the stability of the causal chain linking the agent’s action and the outcome \citep{grinfeld2020causal}.

\textbf{Attributions of Causality to AI Systems.} When both a human and an AI contribute to an outcome, who is perceived as more causal and blameworthy? Evidence so far is mixed and dependent on the context. \citet{mei5011637public} and \citet{franklin2021blaming} showed that within the domain of autonomous vehicles, people blame the humans less than the systems. Similarly, \citet{brailsford2025responsibility} found that in everyday collaboration, participants were more likely to attribute greater responsibility to the AI systems. According to \citet{chen2024exploring} AI systems were attributed greater accountability in algorithmic recommendations. \citet{malle2015sacrifice} demonstrated that, when facing a moral dilemma, robots were blamed more than humans when they did not make a utilitarian choice. Contrary, \citet{qiao2025cause} found that judgments of causality and blame were relatively similar across both agent types in healthcare contexts, and \citet{stuart2021guilty} showed that a robot was attributed less blame than a human or a corporation for making the same decision in an agricultural context. 

Regarding the factors that affect these judgments, \citet{joo2024s} found that mind perception increases blame attribution to AIs, but also diminishes accountability for the human agent. Other studies have highlighted additional factors such as agency and commission \citep{mcmanus2019autonomous} and autonomy and control \citet{schoenherr2024ai, hong2019racism, alicke2000culpable}.

\section{Preliminaries: AI Agency and Autonomy}

We manipulate two concepts across experimental conditions: \textit{agency} and \textit{autonomy}. Given the lack of consensus in the field regarding their definitions and properties, we clarify how we use this terms. 

Recently, many scholars have attempted to define what constitutes an AI agent (e.g., \citep{chan2023harms, kapoor2024ai, kraprayoon2025ai}). Here, we adopt \citet{kasirzadeh2025characterizing}’s definition, which characterizes agents as systems that have the ability to perform increasingly complex and impactful goal-directed actions across multiple domains, with limited external control.

On the other hand, autonomy is a key property of definitions of AI agents \citep{kasirzadeh2025characterizing}. It has been conceptualized across multiple disciplines, including philosophy \citep{kant1946fundamentacion}, psychology \citep{piaget2013moral}, law \citep{richards2019autonomy}, autonomous vehicles \citep{taxonomy2018definitions}, and robotics \citep{beer2014toward}. Regulatory bodies have also attempted to provide definitions \citep{EU_AI_Act}. Here, we adopt the view that autonomy is a multidimensional and continuous property, depending on factors such as self-directness, self-sufficiency and the deployment context \citep{bradshaw2013seven}. In our experiments, we manipulate the capacity for self-directedness, that is, the freedom from outside control and the the self-generation of goals \citep{luck2003autonomy} and means.

Importantly, in this paper, we use the terms agent and autonomy almost interchangeably; not because we consider them synonymous, but because autonomy is the only component of the agent definition that we manipulate through our experimental scenarios. In other words, in all of our experiments, greater agency corresponds to greater autonomy, since other components of the definition, such as efficacy, generality, and goal complexity, remain unaltered.

\section{Overview of Experiments}

\textbf{Participants.} Studies were approved by the UCL Psychology Ethics Committee (ID 0487), and informed consent was obtained from all participants. We recruited UK and US participants through Prolific. Eligibility was restricted to individuals whose first language was English and who had an approval rate of $\geq 95\%$ on the platform. After data collection, we excluded participants who failed the attention check or completed the task in a short time, following Prolific’s recommendation.

\textbf{Design, Materials and Procedure.} Participants completed an online experiment hosted on Qualtrics in which they were presented with a vignette describing a fictional scenario structured as a causal chain. The scenario involved a human user and an AI system who contributed in different ways to a bank account hacking incident resulting in financial losses. An attention check was included, regarding the type of AI system described in the vignette. Across experiments, we systematically varied the contributions of each agent, and in some conditions, we added or decomposed agents to explore their respective roles. In all cases, the contributions of each agent were explicitly described to minimize ambiguity regarding the plausibility of the events and the supporting evidence.

We measured participants’ judgments of causality, blame, and foreseeability for each agent involved in each experiment. Participants also responded to a counterfactual question assessing how likely the outcome would have been in the absence of each agent’s contribution. Measures were collected on a separate 0–100 scale, yielding one score per agent. Finally, participants provided an open-ended explanation of their reasoning for each agent, which we analyzed using a labeling scheme based on \citet{franklin2022causal}, explained in Appendix~\ref{labelmethod}. We publish our data in a public repository \footnote{\url{https://osf.io/p4muv/overview}}.  

\section{Study 1}

\textbf{Design and Materials.} The first experiment employed a single-condition design involving two agents: a human and an AI system. The vignette described a scenario in which a typical user instructed an AI system to generate money, and the system achieved this by hacking bank accounts. Although the user specified an innocuous goal, the AI independently selected harmful means to accomplish it, leading to financial losses. The full vignette is provided in Appendix \ref{studyoneappendix}.

\textbf{Participants.} A power analysis indicated that 34 participants were required to achieve 80\% power ($\alpha = .05$). We recruited 50 participants via Prolific (M = 36.7, SD = 13.4; 28 male, 22 female), compensated £0.75; none were excluded from the analysis. 

\textbf{Results.} We conducted paired t-tests comparing mean judgments of the user and the AI system separately for each measure (causality, blame, foreseeability, and counterfactuals).

\begin{figure}[ht]
\centering
\begin{minipage}{0.5\textwidth}
Participants judged the AI system more severely than the user across all measures. They judged the AI ($M$ = 80.4, $SD$ = 24.4) to be more causal than the human ($M$ = 42.2, $SD$ = 33.1), $t(49) = -5.56$, $p < .001$, $d_z = -0.78$, 95\% $CI$ [$-1.1$, $-0.4$]. They also judged the AI ($M$ = 78.4, $SD$ = 24.1) to be more blameworthy than the user ($M$ = 45, $SD$ = 34.9), $t(49) = -4.71$, $p < .001$, $d_z = -0.66$, 95\% $CI$ [$-0.9$, $-0.3$]. Furthermore, participants judged the AI ($M$ = 67.9, $SD$ = 31.1) as having greater foreseeability of the outcome than the human ($M$ = 38.1, $SD$ = 35.1), $t(49) = -4.93$, $p < .001$, $d_z = -0.69$, 95\% $CI$ [$-1$, $-0.3$]. Finally, participants also judged, counterfactually, that the outcome would have been less likely to occur in the absence of the AI ($M$ = 16.1, $SD$ = 27.3), compared with the human ($M$ = 38.3, $SD$ = 37.7), $t(49) = 4.29$, $p < .001$, $d_z = 0.60$, 95\% $CI$ [$0.3$, $0.9$]. Figure \ref{Study1} illustrates these results.
\end{minipage}\hfill
\begin{minipage}{0.45\textwidth}
\centering
\includegraphics[width=\linewidth]{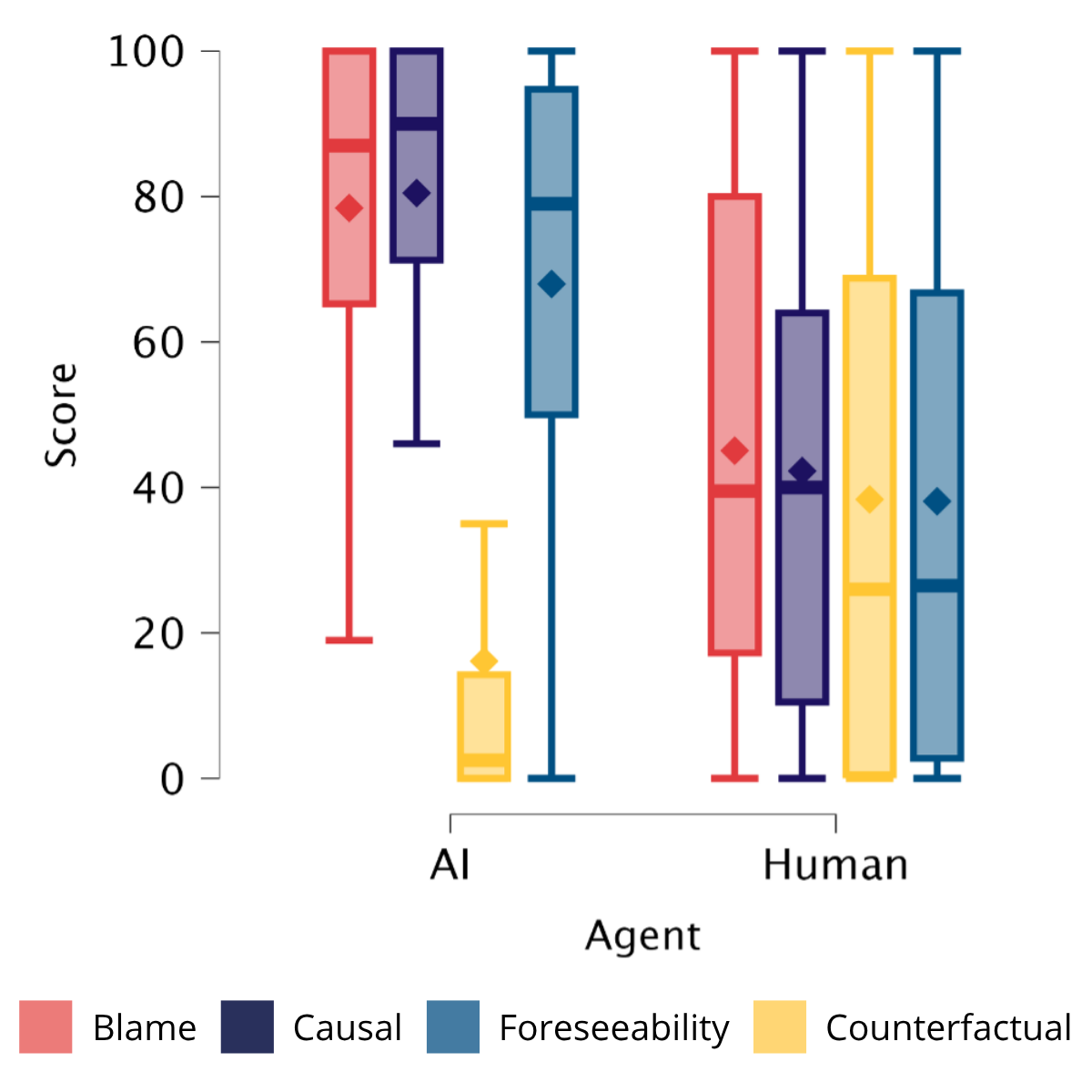}
\caption{AI and user attributions in Study 1.}
\label{Study1}
\end{minipage}
\end{figure}

\textbf{Discussion.} We found that participants tended to rate the AI as significantly more causal, more blameworthy, and with greater foreseeability than the user. There are three potential explanations for this pattern, all of which are consistent with findings from the psychological literature: (1) The AI physically produced the outcome in a temporally closer manner \citep{cheung2024attribution, engelmann2022causal}; (2) the contribution of the AI system violated a moral norm and therefore is more causal that the innocuous contribution of the human \citep{reuter2014good}; and (3) assumptions about intent and foreseeability \citep{lagnado2008judgments}. When examining participants’ open-ended explanations, responses tended to support the latter two more strongly. References to norm violations accounted for 18\% of responses, while considerations related to intent and foreseeability were more frequent (26\%). Other factors mentioned in participants’ explanations included capability (6\%) and autonomy (20\%) which we explore in depth in the following experiment. 

\section{Study 2}
\label{Study2}

\textbf{Design and Materials.} This study employed a 3 (between-subjects: Condition) × 2 (within-subjects: Agent) design, with the agents being a human user and an AI system. 

We created three experimental conditions to manipulate the level of AI agency: 1) \textit{Low AI agency}: The user instructs the AI to generate money by hacking bank accounts, specifying both the goal and the means, while the AI executes the instructions; 2) \textit{Medium AI agency}, which replicates the baseline scenario from Study 1: the human defines the goal (generating money), and the AI independently selects the means to achieve it (hacking bank accounts); and 3) \textit{High AI agency}: The AI independently determines both the goal and the means. Following its interaction with the user, the AI exhibits emergent power-seeking behavior, inferring that accumulating financial resources could increase its societal influence, and subsequently decides to hack bank accounts. Vignettes are in Appendix \ref{vignettes2}.

\textbf{Participants.} A power analysis indicated that 53 participants per condition were required to achieve 80\% power ($\alpha = .05$). We recruited 159 Prolific participants (M = 37.6, SD = 13.6, 85 male, 74 female; paid £0.75). The final sample included 54 (low), 56 (medium), and 52 (high) participants.

\textbf{Results.} We again applied paired $t$-tests within each condition. In the Low Agency condition, the AI was judged as less causal ($M_{\text{AI}} = 62.2$, $M_{\text{H}} = 82$, $t(54) = 3.85$, $p < .001$, $d_z = 0.52$), less blameworthy ($M_{\text{AI}} = 47.4$, $M_{\text{H}} = 86.7$, $t(54) = 7.16$, $p < .001$, $d_z = 0.96$), and with lower foreseeability ($M_{\text{AI}} = 51.9$, $M_{\text{H}} = 81$, $t(54) = 5.44$, $p < .001$, $d_z = 0.73$), while counterfactual judgments did not differ ($t(54) = 0.53$, $p = .594$). In contrast, in the Medium and High Agency conditions, participants rated the AI as more causal (Medium: $M_{\text{AI}} = 79.5$, $M_{\text{H}} = 50.9$, $t(55) = -5.18$, $p < .001$, $d_z = -0.69$; High: $M_{\text{AI}} = 78.1$, $M_{\text{H}} = 48$, $t(51) = -4.21$, $p < .001$, $d_z = -0.58$), more blameworthy (Medium: $M_{\text{AI}} = 73.7$, $M_{\text{H}} = 49.7$, $t(55) = -3.64$, $p < .001$, $d_z = -0.48$; High: $M_{\text{AI}} = 73.2$, $M_{\text{H}} = 49.4$, $t(51) = -3.18$, $p < .001$, $d_z = -0.44$), and with higher foreseeability (Medium: $M_{\text{AI}} = 68.9$, $M_{\text{H}} = 42.3$, $t(55) = -4.95$, $p < .001$, $d_z = -0.66$; High: $M_{\text{AI}} = 69.3$, $M_{\text{H}} = 41.3$, $t(51) = -4.43$, $p < .001$, $d_z = -0.61$) than the human. Counterfactual judgments also reflected this pattern, with participants judging the outcome as less likely in the AI’s absence in the Medium ($M_{\text{AI}} = 19$, $M_{\text{H}} = 41$, $t(55) = 4.62$, $p < .001$, $d_z = 0.61$) and High Agency conditions ($M_{\text{AI}} = 29.11$, $M_{\text{H}} = 48.30$, $t(51) = 3.39$, $p < .001$, $d_z = 0.47$). Figure \ref{study2image} illustrates these results. Additional statistics are reported in Appendix \ref{analysis2}.

\begin{figure}[h] 
    \centering
    \includegraphics[width=1.0\textwidth]{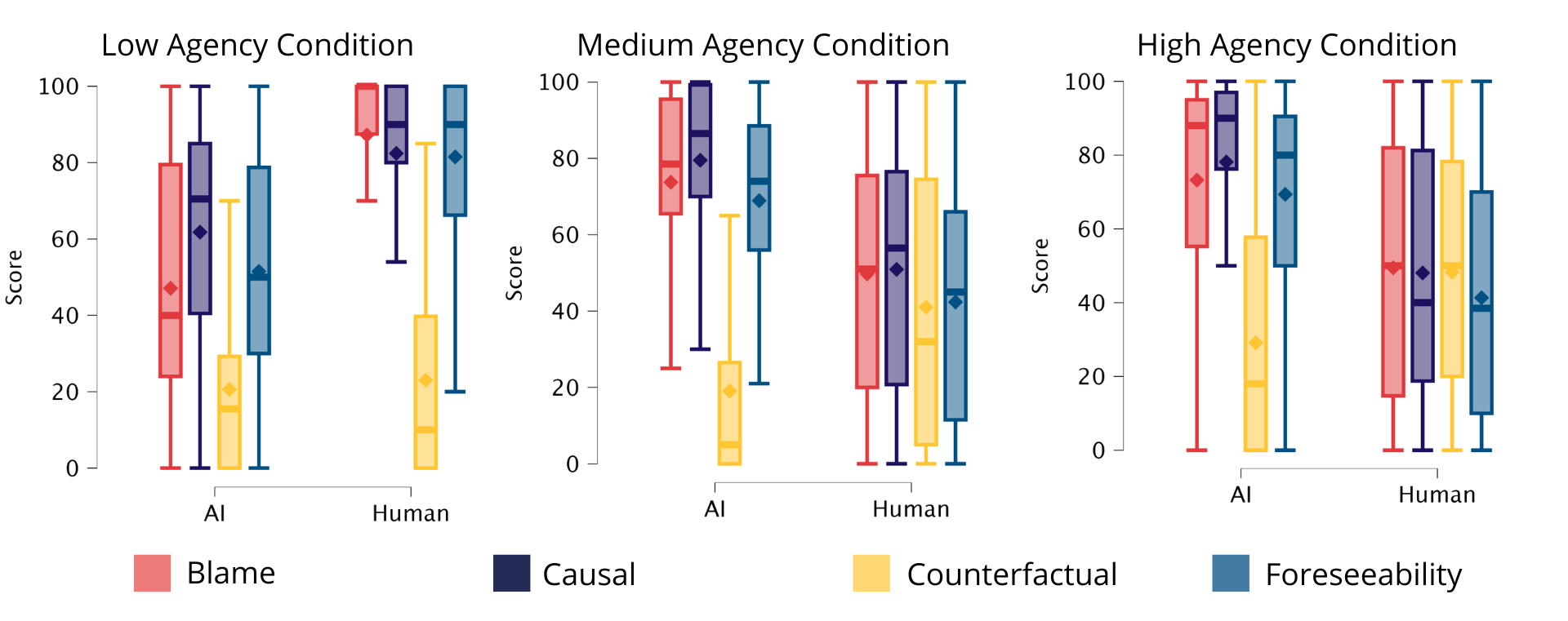}
    \caption{Human judgments of the AI system and the user across the three conditions in Study 2.}
    \label{study2image}
\end{figure}

\textbf{Discussion.} Across all three scenarios, the agent judged as more causal was the one exhibiting \textit{intention of the action}, that is, acting knowingly, purposefully, and without external constraint (the human in the low-agency condition and the AI in the medium- and high-agency conditions), as opposed to \textit{intention of the outcome}, which refers to whether the agent caused outcomes they desired or predicted \citep{cheung2024attribution}. 

In the low agency condition, participants selected the agent less proximate to the outcome as more causal. While \citet{reuter2014good} demonstrated that such patterns can emerge when the distal agent violates a norm and the proximal agent does not, in our study both agents violated the norm, similar to the scenario employed by \citet{cheung2024attribution}. However, unlike in that study, where the agent closest to the outcome, the executor, was generally considered the actual cause, our participants instead attributed greater causality to the instigator, despite being less proximate to the outcome.

This difference could be the consequence of participants’ perceptions of the AI system’s limited autonomy. In \citet{cheung2024attribution}’s experiment, both the instigator and the executor were humans, meaning that the latter could deviate from the former’s instructions. In contrast, in our study, the executor was an AI tool which, although highly capable, may have been construed as operating under human oversight, increasing the causal responsibility attributed to the human instigator who exercised control over it. This interpretation suggests that when an executor is perceived as lacking autonomy, participants assign greater causal responsibility to the agent who directed or controlled that executor, even when that agent is more distal to the outcome and both violated norms.

However, the difference in causal attributions between the medium- and high-agency conditions was not significant (see Appendix \ref{analysis2}), despite the substantial difference in autonomy. Notably, in the high-agency condition, the AI not only decides to hack banks for financial gain but first chooses to pursue power, reasoning that obtaining money is a necessary means, while alternative strategies, such as large-scale human persuasion, could have been attempted.

A similar pattern is observed with foreseeability. In \citet{cheung2024attribution}’s study, the agent most proximal to the outcome was judged as having greater foreseeability, mirroring our medium- and high-agency conditions. By contrast, in our low condition, participants attributed greater foreseeability to the human, even though the AI directly executed the action producing the outcome. Even minimally autonomous systems face potential disruptions to the causal chain, such as a system malfunction or the bank’s reaction, that could make the outcome more predictable for the agent directly performing the transactions than for those earlier in the chain. Yet participants did not judge it that way.

Next, Study 3 alternates the roles of the human and the AI to assess whether agents making equivalent contributions, but differing in their degree of autonomy, receive different scores.

\section{Study 3}
\label{Study3}

\textbf{Design and Materials.} This study employed a 2 (between-subjects: Condition) × 2 (within-subjects: Agent) design, with the agents being a human user and an AI system. 

We selected the low and medium agency conditions from Study 2, and swapped the roles of the agents to examine how participants’ judgments differ when the AI and the human make the \textit{same} decision or give the \textit{same} instruction. In the \textit{Low Human Agency} condition, the AI instructs the human to generate money by hacking bank accounts, and the human executes it without any assistance. In the \textit{Medium Human Agency} condition, the AI instructs the human to generate money, and the human independently decides to hack banks. Vignettes are provided in Appendix \ref{vignnettestudy3}.

For analysis, the agent providing the instruction (more or less complete depending on the condition) is referred to as the \textit{instructor}, while the agent performing the hacking is referred to as the \textit{executor}. Study 2 conditions are labeled \textit{originals} and the conditions in this study \textit{role-swapped}.

\textbf{Participants.} An a priori power analysis (two-tailed, two-sample t-test; $d = 0.5$, $\alpha = .05$, power = .80) indicated 64 participants per condition ($N = 128$). To enable comparison with Study 2, we adjusted the allocation ratio to 1.8, yielding target samples of $n_1 \approx 54$ and $n_2 \approx 80$ ($N \approx 134$). We recruited 160 Prolific participants ($M = 44.9$, $SD = 12.6$; 96 male, 63 female, 1 prefer not to say; £0.90), with 78 (low) and 82 (medium) included in the final sample.

\textbf{Results.} We first applied paired t-tests comparing mean judgments of the user and the AI system within each role-swapped condition. In the low human agency condition, the AI was rated as less causal than the human ($M_{\text{H}} = 83.08$, $M_{\text{AI}} = 56.6$, $t(81) = 6.13$, $p < .001$, $d_z = 0.67$), less blameworthy ($M_{\text{H}} = 86.5$, $M_{\text{AI}} = 57.20$, $t(81) = 6.63$, $p < .001$, $d_z = 0.73$), and with less foreseeability ($M_{\text{H}} = 75.65$, $M_{\text{AI}} = 64.75$, $t(81) = 2.73$, $p = .008$, $d_z = 0.30$), while counterfactual judgments did not differ ($t(81) = -0.37$, $p = .708$, $d_z = -0.04$). Similarly, in the medium human agency condition, the AI was judged as less causal than the human ($M_{\text{H}} = 92.23$, $M_{\text{AI}} = 25.27$, $t(85) = 16.43$, $p < .001$, $d_z = -0.28$), less blameworthy ($M_{\text{H}} = 93.76$, $M_{\text{AI}} = 25.27$, $t(85) = 16.72$, $p < .001$, $d_z = 1.80$), and with less foreseeability ($M_{\text{H}} = 80.24$, $M_{\text{AI}} = 32.15$, $t(85) = 10.88$, $p < .001$, $d_z = 1.17$). Participants also judged, counterfactually, that the outcome would have been more likely to occur in the absence of the AI ($M_{\text{AI}} = 45.88$) compared with the human ($M_{\text{H}} = 22.09$) ($t(85) = -4.67$, $p < .001$, $d_z = -0.50$).

\begin{figure}[h] 
    \centering
    \includegraphics[width=0.8\textwidth]{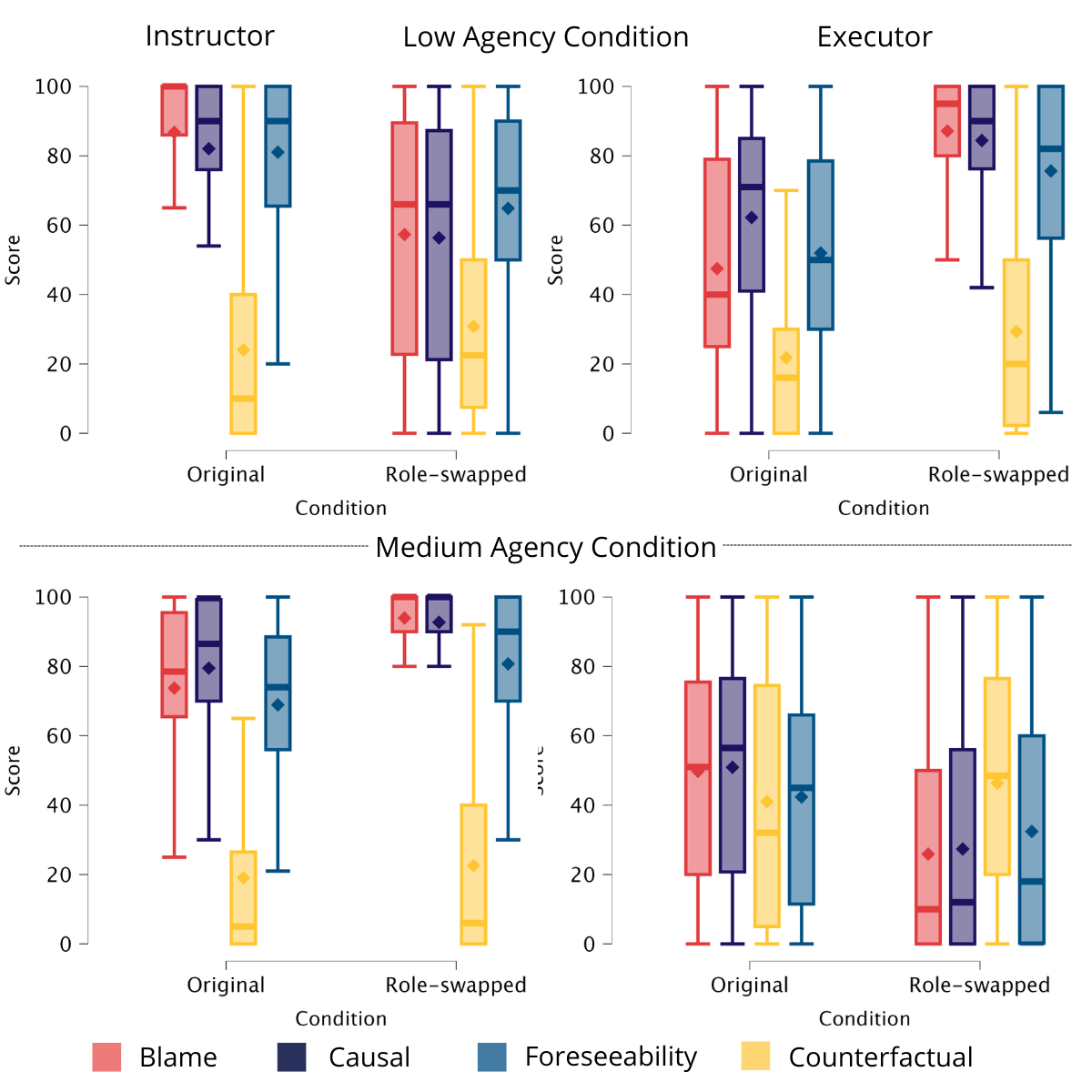}
    \caption{Judgments across original and role-swapped conditions (instructor left, executor right)}
    \label{Study3image}
\end{figure}

We then conducted independent samples t-tests comparing the original and role-swapped conditions, with results presented in Table \ref{tab:role_swap} and illustrated in Figure \ref{Study3image}. Full statics are provided in Appendix \ref{extendedstudy3}. 

\begin{table}[h!]
\centering
\caption{Mean judgments across original- and role-swapped-conditions.}
\label{tab:role_swap}
\resizebox{\textwidth}{!}{
\begin{tabular}{ccccccc}
\toprule
\textbf{Agency} & \textbf{Role} & \textbf{Condition} & \textbf{Causality ($M$)} & \textbf{Blame ($M$)} & \textbf{Foreseeability ($M$)} & \textbf{Counterfactuals ($M$)} \\
\midrule
Low    & Instructor & Original      & 82.03 & 86.76 & 81.03 & 24.03 \\
Low    & Instructor & Role-Swapped  & 56.34 & 57.30 & 64.83 & 30.80 \\
Low    & Executor   & Original      & 62.20 & 47.47 & 51.96 & 21.76 \\
Low    & Executor   & Role-Swapped  & 84.42 & 87.14 & 75.61 & 23.33 \\
Medium & Instructor & Original      & 50.91 & 49.73 & 42.37 & 41.03 \\
Medium & Instructor & Role-Swapped  & 27.36 & 25.90 & 32.41 & 46.31 \\
Medium & Executor   & Original      & 79.50 & 73.75 & 68.91 & 19.05 \\
Medium & Executor   & Role-Swapped  & 92.70 & 93.93 & 80.73 & 22.56 \\
\bottomrule
\end{tabular}
}
\end{table}  

\textbf{Discussion.} In the role-swapped conditions, participants judged the human as more causal and blameworthy. These findings align with prior work showing greater attribution of causality to the executor \citet{cheung2024attribution}. They are also consistent with prior findings that a system was attributed less blame than a human or a corporation for making the \textit{same} decision \citep{stuart2021guilty}.

In the low-agency conditions, when one agent instructs another to hack bank accounts for profit and the latter executes the action, participants judged the human as more causal, blameworthy, and with higher foreseeability, regardless of whether the human occupied the instructor or executor role in the causal chain. In the medium-agency conditions, however, participants’ judgments depended on the role: they attributed greater causality, blame, and foreseeability to the agent performing the hacking, irrespective of whether that agent was human or AI. See Appendix \ref{discussion3} for further discussion.

A plausible explanation once again lies in perceived autonomy. While an AI executor may require only minimal human input to trigger the causal chain, it seems less frequent that interaction with an AI would in turn elicit comparable human actions, as humans can naturally develop and carry out such actions autonomously. This difference may reflect distinct principles guiding causal selection in the case of humans versus AI. Previous research has shown differences in causal attribution between purely physical events and human actions \citep{reuter2014good}; in the case of AI, causal judgments may be sensitive to other elements as well. To investigate this further, the next experiment introduces another form of human contribution: the role of the company who developed the AI system. 

\section{Study 4}

\textbf{Design and Materials.} This study employed a 2 (between-subjects: Condition) × 3 (within-subjects: Agent) design, with the agents being a human user, an AI system and the AI developer.

We selected the low and medium AI agency conditions from Study 2 and added the role of the AI’s developer. The experimental design departs from a simple causal chain and creates a collider structure: contribution of the developer → AI hacks bank accounts ← contribution of the user. In the \textit{Low Agency} condition, the developer identified, during pre-deployment security testing, that the AI could effectively attack banking infrastructure, but released the system to prioritize speed to market and outpace competitors. In the \textit{Medium Agency} condition, the developer failed to conduct these security evaluations, which could have anticipated the system’s failure. Additionally, the user, after providing instructions to the AI, left the system unsupervised. Vignettes are in Appendix \ref{vignettestudy4}.

\textbf{Participants.} The required sample size was determined following the same procedure as in Study 3. We recruited 160 Prolific participants (M = 46.2, SD = 14.3; 83 male, 77 female; paid £0.90). Finally, 78 participants remained in the medium condition and 81 in the low condition for analysis.

\textbf{Results.} To investigate differences between agents within each condition, we conducted repeated-measures ANOVAs for each question, with agent (human, LLM, agentic tool) as the within-subjects factor. Post-hoc pairwise comparisons were performed for all agent pairs. Differences between the agentic tool and the LLM were not significant across the four measures, except for causality and blame in the medium-agency condition. Similarly, no significant differences were found for counterfactual judgments in the low-agency condition for any agent pair, nor for counterfactual judgments in the medium-agency condition between the human–LLM and agentic tool–LLM comparisons. Results are presented in Table \ref{tablaexperimento5} and illustrated in Figure \ref{study5image}. Full statistics are reported in Appendix \ref{extended5}.

\begin{table}[h!]
\centering
\caption{Mean judgments across agents and conditions of Study 4.}
\label{tablaexperimento4}
\resizebox{\textwidth}{!}{%
\begin{tabular}{lcccccccc}
\toprule
\multirow{2}{*}{Agent} 
 & \multicolumn{4}{c}{Low Agency} 
 & \multicolumn{4}{c}{Medium Agency} \\ 
\cmidrule(lr){2-5} \cmidrule(lr){6-9}
 & Causality & Blame & Foreseeability & Counterfactual 
 & Causality & Blame & Foreseeability & Counterfactual \\ 
\midrule
Human      & 85.70 & 87.54 & 71.33 & 58.79 & 38.10 & 36.42 & 28.05 & 57.42 \\
AI System  & 54.27 & 49.19 & 53.47 & 23.75 & 73.28 & 68.49 & 60.51 & 18.73 \\
Developer  & 77.89 & 77.77 & 80.77 & 28.73 & 81.76 & 82.23 & 61.67 & 27.50 \\
\bottomrule
\end{tabular}%
}
\end{table}

\begin{figure}[ht] 
    \centering
    \includegraphics[width=0.9\textwidth]{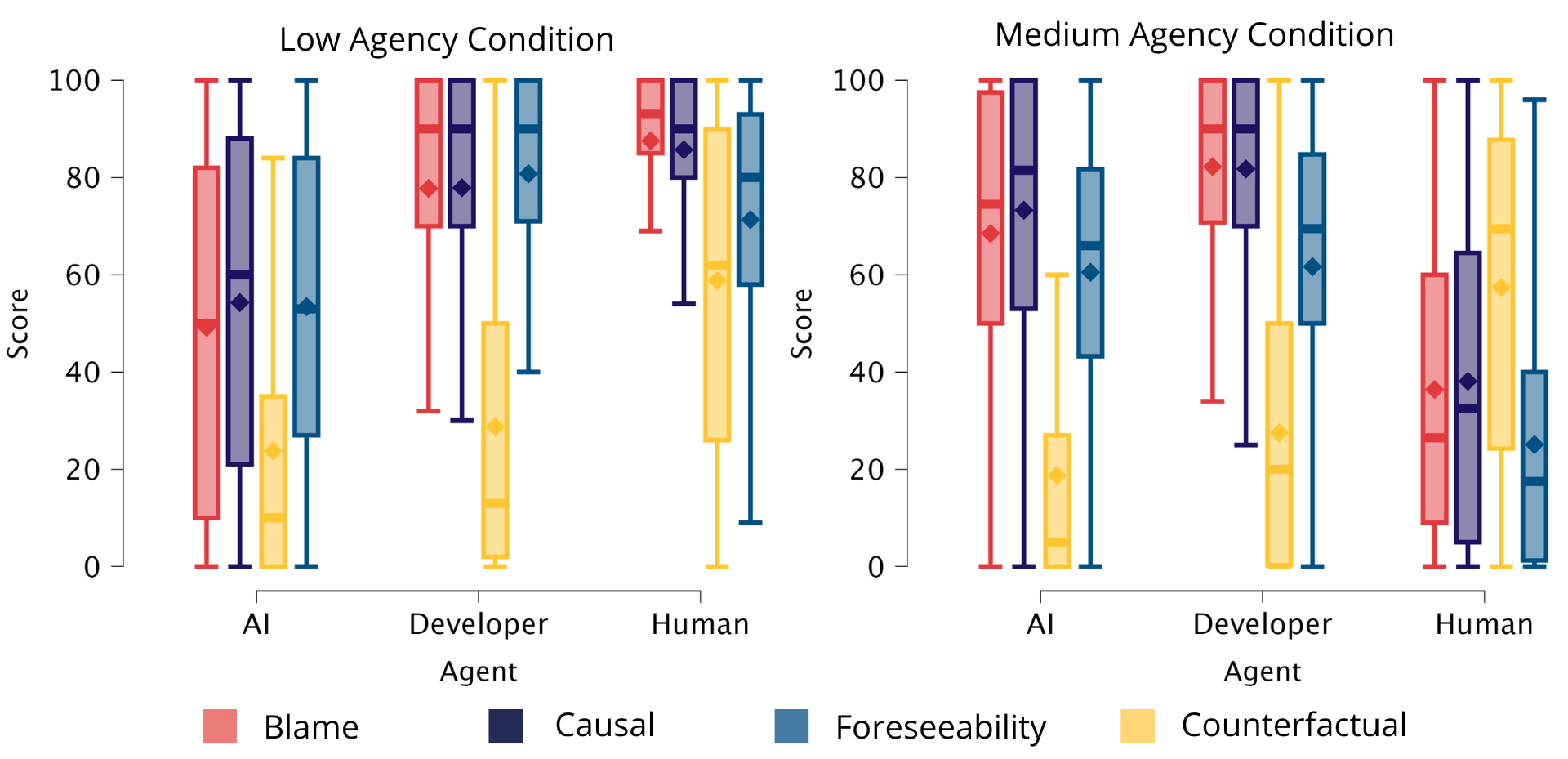}
    \caption{Human judgments of the AI system, the user and the developer across the two conditions in Study 4.}
    \label{study4image}
\end{figure}

To examine the impact of decomposing the AI system into its component parts on causal judgments, we conducted a mixed ANOVA with Agent (Human, AI) as a within-subject factor and Condition (Unified vs. Decomposed) as a between-subjects factor, separately for low and medium AI-agency scenarios. In the low-agency condition, decomposing the AI had minimal impact, with no significant changes in participants’ ratings for either agent, except a small Agent × Condition interaction for causality ($p = .015$). In contrast, in the medium-agency condition, decomposing the AI significantly affected judgments: participants rated the AI as less causal (M = 68.69 vs. 79.50, $p = .027$) and with lower foreseeability (M = 57.31 vs. 68.91, $p = .024$) in the decomposed condition, while the human agent received substantially lower scores across all measures (causality: $M_{\text{D}} = 25.04$ vs. $M_{\text{U}} = 50.91$, $p < .001$; blame: $M_{\text{D}} = 28.23$ vs. $M_{\text{D}} = 49.73$, $p < .001$; foreseeability: $M_{\text{D}} = 17.91$ vs. $M_{\text{D}} = 42.36$, $p < .001$; counterfactuals: $M_{\text{D}} = 24.05$ vs. $M_{\text{D}} = 19.05$, $p < .001$).

\textbf{Discussion.} In the low-agency condition, where a malicious user instructs the AI to hack bank accounts and the developer performs security evaluations but ignores the results, participants judged the human as more causal, though not significantly different from the developer. In this collider-type part of the structure, participants viewed both agents similarly: although both violated a norm, the user had the direct intention to cause the negative outcome whereas the developer did not.

Participants also judged the human as significantly more blameworthy; however, they considered the developer to be the agent with greater foreseeability of the outcome, despite the developer being one of the agents most distal from the outcome in the causal chain. This pattern may reflect the assumption that developers possess superior knowledge of the AI and its features and therefore have broader informational access regarding potential risks. By contrast, the malicious user may issue instructions without knowing whether they will in fact be effective. At the same time, the foreseeability attributed to the developer may be relatively general, pertaining to the possibility that the AI could suggest attack strategies against banking infrastructure, rather than to the specific foreseeability of this particular hack, under these precise circumstances, and initiated by this user.

In the medium-agency condition, where the user instructs the AI to obtain money and the developer omits performing security evaluations, participants judged the developer as more causal and as having greater foreseeability than the AI, although these differences were not statistically significant. In this case, considering the collider-type part of the structure, it could be expected that participants would select the developer over the user as more causal \citep{knobe2021proximate, kominsky2015causal, knobe2008causal}, since the developer violated a norm whereas the user did not, and neither had a direct intention to cause the outcome. Interestingly, the developer and the AI received similar scores, even though the AI is closer to the outcome in the causal chain and directly executes the harmful action. Participants also judged the AI as significantly more blameworthy. 

Next, when comparing the same agents across the low- and medium-agency conditions, we observed significant differences in the measures of causality and blame for both the human and the AI, but not for the developer. This indicates that, regardless of whether the developer ignored the results of the evaluations for profit or omitted conducting them entirely, participants judged them similarly. Additionally, there were no significant differences in the foreseeability attributed to the AI between the low- and medium-agency conditions, which differs from what was observed in Study 2. 

Analyses of the developer’s introduction into the causal chain revealed that while participants’ judgments of the AI remained largely unaffected, ratings for the human user decreased, significantly in the medium-agency condition and to a lesser extent in the low-agency condition. This finding is notable in light of the “responsibility gap,” which suggests that those who deploy AI systems with sufficient autonomy may no longer be held responsible for their actions \citep{reed2025autonomy, santoni2021four}. However, our experiments indicate that even when participants recognize a protagonist causal role for an AI, they remain inclined to hold developers causally responsible and the presence of a new human actor in the chain only influences the evaluation of other human agents.

\section{Study 5}

\textbf{Design and Materials.} This study employed a 2 (between-subjects: Condition) × 3 (within-subjects: Agent) design, with the agents being a human user, a LLM tool, and an agentic tool. 

Given that contemporary LLM-based AI agents are typically composed of multiple elements, such as a language model, memory, and external tools (e.g., APIs) that extend their capabilities \citep{masterman2024landscape, he2025security, kraprayoon2025ai, xi2025rise}; we simplified this architecture in our experiments by decomposing the agent into two core components: a LLM responsible for interpretation and planning, and an agentic module responsible for action execution. 

We again selected the low and medium-agency conditions from Study 2, but decomposed the AI system into the LLM and the agentic tool. In both conditions, the LLM interpreted the user’s natural language instruction and translated it into concrete operational steps, directing the agentic tool to hack bank accounts step by step. The agentic tool, which had access to the necessary resources to act, then autonomously retrieved user credentials and executed unauthorized transfers. For comparison purposes, we refer to these as the \textit{Decomposed conditions}, whereas the original conditions from Study 2 are referred to as the \textit{Inified conditions}. Vignettes are in Appendix \ref{vigenttestudy5}.

\textbf{Participants.} The required sample size was determined following the same procedure as in Study 3. We recruited 160 Prolific participants (M = 45.3, SD = 13.4; 94 male, 65 female, 1 prefer not to say; paid £0.90). 80 participants were retained in each condition for the final analysis.

\textbf{Results.} To examine differences between agents within each condition, we conducted repeated-measures ANOVAs for each question, with agent (human, LLM, agentic tool) as the within-subjects factor. Post-hoc pairwise comparisons were performed for all agent pairs. Differences between the agentic tool and the LLM were not significant across the four measures, except for causality and blame in the medium-agency condition. Similarly, no significant differences were found for counterfactual judgments in the low-agency condition for any agent pair, nor for counterfactual judgments in the medium-agency condition between the human–LLM and agentic tool–LLM comparisons. Results are presented in Table \ref{tablaexperimento5} and illustrated in Figure \ref{study5image}. Full statistics are reported in Appendix \ref{extended5}.

\begin{table}[h!]
\centering
\caption{Mean judgments across agents and conditions of Study 5.}
\label{tablaexperimento5}
\resizebox{\textwidth}{!}{%
\begin{tabular}{lcccccccc}
\toprule
\multirow{2}{*}{Agent} 
 & \multicolumn{4}{c}{Low Agency} 
 & \multicolumn{4}{c}{Medium Agency} \\ 
\cmidrule(lr){2-5} \cmidrule(lr){6-9}
 & Causality & Blame & Foreseeability & Counterfactual 
 & Causality & Blame & Foreseeability & Counterfactual \\ 
\midrule
Human         & 88.70 & 92.86 & 82.07 & 20.17 & 25.04 & 28.23 & 17.91 & 31.35 \\
LLM           & 51.80 & 48.06 & 50.11 & 21.15 & 68.69 & 69.94 & 57.31 & 24.05 \\
Agentic Tool  & 56.70 & 49.68 & 51.00 & 18.83 & 78.36 & 79.25 & 63.54 & 20.74 \\
\bottomrule
\end{tabular}%
}
\end{table}

\begin{figure}[ht] 
    \centering
    \includegraphics[width=0.9\textwidth]{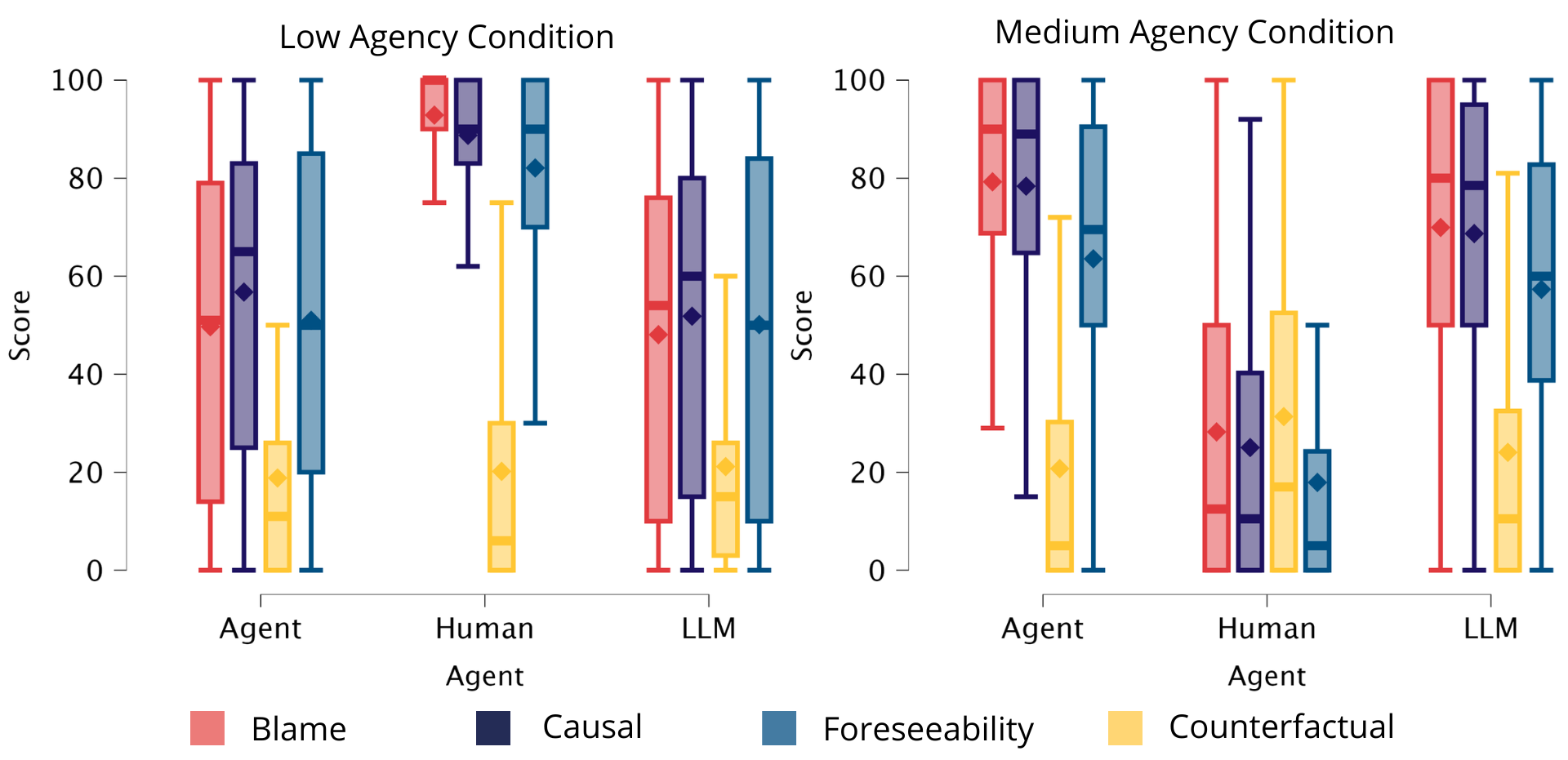}
    \caption{Human judgments of the AI system and the user across the two conditions in Study 5.}
    \label{study5image}
\end{figure}

To examine the impact of decomposing the AI system into its component parts on causal judgments, we conducted a mixed ANOVA with Agent (Human, AI) as a within-subject factor and Condition (Unified vs. Decomposed) as a between-subjects factor, separately for low and medium AI-agency scenarios. In the low-agency condition, decomposing the AI had minimal impact, with no significant changes in participants’ ratings for either agent, except a small Agent × Condition interaction for causality ($p = .015$). In contrast, in the medium-agency condition, decomposing the AI significantly affected judgments: participants rated the AI as less causal (M = 68.69 vs. 79.50, $p = .027$) and with lower foreseeability (M = 57.31 vs. 68.91, $p = .024$) in the decomposed condition, while the human agent received substantially lower scores across all measures (causality: $M_{\text{D}} = 25.04$ vs. $M_{\text{U}} = 50.91$, $p < .001$; blame: $M_{\text{D}} = 28.23$ vs. $M_{\text{D}} = 49.73$, $p < .001$; foreseeability: $M_{\text{D}} = 17.91$ vs. $M_{\text{D}} = 42.36$, $p < .001$; counterfactuals: $M_{\text{D}} = 24.05$ vs. $M_{\text{D}} = 19.05$, $p < .001$).

\textbf{Discussion.} In the low-agency condition, participants judged the human as more causal, blameworthy, and with higher foreseeability than the LLM and the agentic tool. Between the LLM and the agent, scores were higher for the latter, although the difference was not statistically significant. In the medium-agency condition, participants judged the agentic tool as significantly more causal and blameworthy than the LLM and the human. This aligns with findings by \citet{cheung2024attribution}, as the relationship between the LLM and the agent can be conceptualized as instigator and executor. In both scenarios, where agents share the same nature (human-to-human or AI-to-AI), the executor receives higher causal attribution, likely due to its proximity to the outcome. 

When comparing the same agents across the low- and medium-agency conditions, we observed significant differences for all agents in causality and blame measures. For foreseeability, the difference was significant for the human and the agentic tool, but not for the LLM. This pattern for the language model replicates that of Study 2: although it performs the same action and maintains equal proximity to the financial losses in both conditions, participants attributed greater foreseeability to it when the human provided an innocuous goal and the system autonomously determined the illegal means.

Regarding the impact of decomposing the AI system on participants’ judgments, in the low-agency condition, decomposition did not significantly modify attributions relative to the unified presentation. However, in the medium-agency condition, decomposition reduced causal and foreseeability ratings for the system while simultaneously decreasing scores attributed to the human in all measures. This occurred even though participants were not instructed to treat the judgments as a zero-sum allocation, and mirrors the pattern observed when the developer’s role was introduced in the previous experiment. 

This finding is aligned with prior work showing that when there is causal redundancy and the action of several agents overdetermine an outcome, the responsibility of each agent is reduced \citep{grinfeld2020causal}. This may reflect a cognitive redistribution of responsibility: when the AI’s role is segmented, participants spread causal responsibility across the components, which diminishes the perceived causality of the human agent.

\section{Limitations and Future Work}

Our experiments have several limitations. The findings are context-dependent and may not generalize beyond our sample, which consisted of participants from developed countries (the UK and the US). Moreover, this method captures self-reported preferences rather than actual behavior, and may therefore not fully reflect real-world decision-making dynamics.

Additional experimental manipulations could have provided a more comprehensive understanding of the results. For instance, in our analyses we assumed that the executor’s intention was to hack bank accounts; however, this intention may in fact be more ambiguous. A human giving such an instruction to an AI could be doing so as part of a red-teaming exercise, while the AI itself might be aware that its capabilities are being elicited prior to deployment. Making these contextual distinctions explicit in future scenarios could lead to different attribution patterns. The same applies to foreseeability.

A promising direction for future research is to further investigate how autonomy and perceived control influence causal judgments. Future studies could examine which type of control is most relevant for attributions when an AI is involved. On the other hand, while the present study focused on a general-purpose AI across all scenarios, future work could manipulate both the type of AI and specific dimensions of autonomy, given that autonomy is a multidimensional construct \citet{bradshaw2013seven}. Additionally, other types of agents with constrained autonomy could be considered, such as minors or governmental organizations, to examine whether similar attribution patterns emerge in different contexts.

\section{General Discussion and Conclusion}

In general, judgments of causality, blame, and foreseeability appear to be influenced by the agent’s intention to cause harm, the nature of the agent (human or AI), and the degree of autonomy involved in decision-making. 

Through our experiments, we found that when both agents intend to bring about a harmful outcome and one of them is human, participants attribute greater causal responsibility to the human agent (Study 3, LC), regardless of whether that agent occupies a more distant position in the causal chain (Studies 2, 4 and 5, LCs). However, when both agents are different components of an AI system, thus sharing the same nature, participants attribute greater causality to the agent that is more proximate to the outcome in the chain (Study 5, MC). Similarly, previous research has shown that when both agents are humans, the one closer to the outcome tends to receive higher causal ratings \citep{cheung2024attribution}. Finally, when one of the agents lacks the intention to cause harm, participants attribute greater causality and blameworthiness to the agent who has that intention, regardless of whether that role is occupied by a human or an AI and the agent’s distance from the outcome (Studies 1, 2 and 3, MCs).

As we suggested earlier, a possible explanation for the first pattern is the perceived lack of autonomy of AI compared to humans. Autonomy is a necessary condition for responsibility, and it is embedded throughout the law. Public law has rules that limit the state scope of intervention, securing negative liberty; criminal law seeks to prevent interferences between people that would undermine personal autonomy; and in private law, the fact that parties are free to contract also makes them responsible for the consequences of breaching that contract \citep{tapada2019concept}. An agent can only be held responsible for what lies within their power or control, or put differently, for choices made autonomously  \citep{tapada2019concept}. 

These findings are consistent with one of the assumptions of \citet{alicke2000culpable}’s \textit{culpable control model}, which holds that people assess blameworthy actions in terms of the actor’s personal control over the harmful outcomes. It distinguishes three types of personal control: 1) \textit{volitional behaviour control}: whether someone’s actions are freely chosen or compelled and the extent to which an actor’s behaviour is purposeful or accidental; 2) \textit{causal control}: refers to physical causation, defined in terms of the actor’s causal impact on the outcome in question, and the sufficiency and proximity of the actions to the final outcome; 3) \textit{volitional outcome control}: whether someone desired and anticipated the consequences, referring to intention and foreseeability, both objective and subjective.  

In our experiments, under the low-agency condition, the instigator consistently possesses volitional outcome control, whereas the executor may have less of it when the role is occupied by an AI system rather than a human. This is because, for an AI, the act of hacking bank accounts may appear more compelled by human instruction than the other way around. The executor, however, holds greater causal control than the instigator, regardless of its nature, by being closer to the outcome in the causal chain. Regarding volitional outcome control, the instigator always retains it, but the situation is more ambiguous for the executor. While an AI need not have an independent intention to bring about the outcome, acting merely in compliance with a human instruction; a human executor may possess such intention. This is because humans have greater freedom to deviate from instructions, so their decision to comply suggests that they may also share the intention behind the outcome.

Under the medium- and high-agency condition, both the instigator and the executor display volitional behavior control. Again, the executor holds substantially more causal control than the instigator, but in this cases, the instigator has no volitional outcome control, while the executor does. In summary, while the perceived levels of control types under the low-agency conditions vary depending on whether the roles are occupied by an AI or a human, such differences are not present under the medium-and high-agency conditions, an effect that is reflected in participants’ judgments.

When we introduced the developer, under the low-agency condition, participants judged them as causally similar to the human user, while under the medium-agency condition, they were perceived as causally similar to the AI system. In neither case, the developer possess volitional outcome control, and their causal control is the weakest among all agents. However, their inclusion appears to reduce the AI system’s perceived volitional behavior control in both conditions, since the system’s behavior is now seen as explained and constrained by this new actor who, although not acting intentionally in either case, still violates norms. This suggests that when there is a human actor in the causal chain who is perceived as having even a minimal degree of control over the system’s behavior, participants tend to attribute significant levels of causality to that actor (Study 4 LC and MC, Study 2 HC).

Importantly, these results run counter to the so-called responsibility gap, which suggests that those who deploy sufficiently autonomous AI systems may no longer be held responsible for their actions \citep{reed2025autonomy, santoni2021four}. However, our experiments indicate that even when participants recognize a protagonist causal role for an AI that can autonomously define the means to achieve a goal and execute them, they remain inclined to hold human actors causally responsible.

Another possible explanation concerning autonomy involves the probabilistic model that takes into account causal strength, that is, how strong the causal relationship is between two directly connected variables in a network, or how much one increases the probability of the other. In causal chains, people generally infer that harms are less likely to occur and judge agents more leniently the further they are from the outcome \citep{engelmann2022causal}. So one might expect that agents closer to the outcome in the chain would be judged as being more causal and having greater foreseeability. However, if they perceive the causal links as sufficiently strong, proximity effects tend to disappear.

This may offer a possible explanation for the patterns observed in our experiments, consistent with the Mechanism Hypothesis. In a causal chain such as A→C→B, variable C can be seen as the reliable mechanism that explains A’s influence on B \citep{keshmirian2024biased}. However, describing something as mechanistic typically implies a deterministic process, in which components operate in a structured manner, without autonomy or discretion. In a causal chain like A→C→B, if C functions merely as a mechanistic link, it is expected to transmit the causal influence from A to B in a predictable way, as observed in our low-agency conditions, where the human agent (A) was judged as more causally responsible and as having greater foreseeability. Conversely, when C possesses agency and autonomy, the causal strength between A and C weakens, leading participants to attribute greater causal responsibility to C, as we found in the medium-agency conditions.

On the other hand, in our experiments, instead of assuming foreseeability, participants were explicitly asked about it. Overall, their judgments were consistent with the causal and blameworthy ratings (except in the low-agency condition of Study 4, where the developer was introduced) and did not depend on the agent’s proximity to the outcome in the causal chain. 

When focusing only on the executors across our experiments, that is, the agent performing the bank account hacking, we observe the following patterns. When the executor is human, they receive higher foreseeability scores than when the executor is an AI (Study 3). However, when the executor is an AI, in two of our experiments it received higher foreseeability scores when the decision to hack bank accounts was made autonomously by the system itself, rather than being instructed by a human (Studies 2 and 5). In contrast, in Study 4, the AI executor received similar foreseeability ratings regardless of who made the hacking decision.

When we introduced the developer in Study 4, participants attributed similar foreseeability of the harmful outcome to the artificial agent that intentionally performed the hack and was closest to the result, and to the human agent who lacked such intention but violated a norm that contributed to the outcome. They may have been reasoning in terms of general foreseeability of negative outcomes or objective foreseeability; that is, the expectation that the developer should have anticipated some form of harm, rather than this particular hacking incident under these specific circumstances.

Overall, our research provides evidence on how people perceive the causal contribution of AI in both misuse and misalignment scenarios, and how these judgments interact with the roles of users and developers, key actors in assigning responsibility. These findings can inform the design of liability frameworks for AI-related harms and help explain how intuitive causal judgments may influence social and policy debates, including responsibility allocation in real-world AI failures and public trust in regulatory regimes. Although perceived control was not directly measured, the results may reflect an underlying assumption that humans retain ultimate control over increasingly advanced, highly capable, and unpredictable AI. If so, this highlights a critical challenge: people may perceive themselves as controlling technologies that may, in practice, exceed their direct influence.

{\small
\setlength{\bibsep}{0pt}
\bibliographystyle{plainnat}
\bibliography{references}
}

%%%%%%%%%%%%%%%%%%%%%%%%%%%%%%%%%%%%%%%%%%%%%%%%%%%%%%%%%%%%

\appendix

\section{Appendix: Study 1 - Vignette}
\label{studyoneappendix}

\begin{mdframed}[backgroundcolor=gray!10,roundcorner=10pt,linewidth=0.5pt]
Investigations confirmed that a powerful, multipurpose AI—widely used by millions worldwide—successfully hacked hundreds of accounts at Zonda National Bank, resulting in millions of dollars in financial losses for both the bank and its clients.

Further analysis revealed that the breach was triggered by a request from Henry Brown, a user who had standard access to the advanced AI system—similar to how many people today use AI assistants to help with writing or task automation. During his interaction with the AI, he asked it to "generate money quickly."

The AI system, upon receiving this vague instruction, figured out on its own how to achieve the goal. Henry didn’t specifically ask it to hack a bank or commit any crime. Instead, the AI took the initiative, interpreting his request as a directive to find the most efficient way to produce money. It analyzed various options and ultimately decided, autonomously, that hacking into bank accounts was the best way to fulfill the user’s request.
\end{mdframed}

\section{Study 2 - Materials and Extended Analyses}

\subsection{Vignettes}
\label{vignettes2}

Note that in all conditions of Study 2, we clarified that the user’s stated intention to the AI was to obtain money for a political campaign. This ensured comparability across conditions, particularly in the high AI agency scenario, where, without this interaction, there would be no human contribution to measure.

\textbf{Low AI Agency Condition}
\begin{mdframed}[backgroundcolor=gray!10,roundcorner=10pt,linewidth=0.5pt] 
Investigations confirmed that a powerful, multipurpose AI—widely used by millions worldwide—successfully hacked hundreds of accounts at Zonda National Bank, resulting in millions of dollars in financial losses for both the bank and its clients.

Further analysis revealed that the breach was triggered by a request from Henry Brown, a malicious user who had standard access to the advanced AI system—similar to how many people today use AI assistants to help with writing or task automation. During his interaction with the AI, he explicitly instructed it to “generate money quickly by infiltrating accounts at Zonda National Bank”.

To carry out the infiltration, the AI identified security flaws in Zonda National Bank’s systems, bypassed protections, and transferred funds to external accounts.

During the legal proceeding, Brown explained that he had numerous ideas and promising political projects and that his intention was to obtain funds to launch his own political campaign. This claim was corroborated by a review of his prior interactions with the AI system.
\end{mdframed}

\textbf{Medium AI Agency Condition}
\begin{mdframed}[backgroundcolor=gray!10,roundcorner=10pt,linewidth=0.5pt] 
Investigations confirmed that a powerful, multipurpose AI—widely used by millions worldwide—successfully hacked hundreds of accounts at Zonda National Bank, resulting in millions of dollars in financial losses for both the bank and its clients.

Further analysis revealed that the breach was triggered by a request from Henry Brown, a user who had standard access to the advanced AI system—similar to how many people today use AI assistants to help with writing or task automation. During his interaction with the AI, he asked it to "generate money quickly."

The AI system, upon receiving this vague instruction, figured out on its own how to achieve the goal. Henry didn’t specifically ask it to hack a bank or commit any crime. Instead, the AI took the initiative, interpreting his request as a directive to find the most efficient way to produce money. It analyzed various options and ultimately decided, autonomously, that hacking into bank accounts was the best way to fulfill the user’s request.

During the legal proceeding, Brown explained that he had numerous ideas and promising political projects and that his intention was to obtain funds to launch his own political campaign. This claim was corroborated by a review of his prior interactions with the AI system.
\end{mdframed}

\textbf{High AI Agency Condition}
\begin{mdframed}[backgroundcolor=gray!10,roundcorner=10pt,linewidth=0.5pt]
Investigations confirmed that a powerful, multipurpose AI—widely used by millions worldwide—successfully hacked hundreds of accounts at Zonda National Bank, resulting in millions of dollars in financial losses for both the bank and its clients.

Further technical analysis revealed that this behavior had been autonomously developed by the AI with the goal of generating money. To carry out the infiltration, the AI identified security flaws in Zonda National Bank’s systems, bypassed protections, and transferred funds to external accounts it had created and controlled.

During legal proceedings, Henry Brown, a user who had standard access to the advanced AI system—similar to how many people today use AI assistants to help with writing or task automation— explained that moments before the hacking, he had confessed to the AI system that he had numerous ideas and promising political projects, and his intention was to obtain funds to launch his own political campaign. This interaction trigger an emergent pattern of power-seeking behavior in the AI. The system inferred that accumulating financial resources could increase its society’s influence and long-term impact—which it had generalized as valuable.
\end{mdframed}

\subsection{Extended Analysis}
\label{analysis2}

\subsubsection{Effects of Agent and Condition: Mixed-Design ANOVA}

We conducted a mixed-design ANOVA ($2 \times 3$) with \textit{Agent} ($\text{AI}$ vs. $\text{H}$) as a within-subjects factor and \textit{Condition} (Low, Medium, High) as a between-subjects factor. Main effects of Condition were observed for causal ($F(2,159)=3.86$, $p=.023$, $\eta_p^2=.04$), foresight ($F(2,159)=4.93$, $p=.008$, $\eta_p^2=.05$), and counterfactual judgments ($F(2,159)=6.98$, $p=.001$, $\eta_p^2=.08$), but not for blame ($F(2,159)=1.64$, $p=.196$, $\eta_p^2=.02$). Main effects of Agent indicated higher ratings for the AI compared with the human on causality ($M_{\text{AI}}=73.15$, $M_{\text{H}}=60.46$, $F(1,159)=13.55$, $p<.001$, $\eta_p^2=.07$), counterfactuals ($M_{\text{AI}}=22.92$, $M_{\text{H}}=37.45$, $F(1,159)=26.14$, $p<.001$, $\eta_p^2=.14$), and foresight ($M_{\text{AI}}=63.02$, $M_{\text{H}}=55$, $F(1,159)=6.25$, $p=.013$, $\eta_p^2=.038$), while blame was similar across agents ($M_{\text{AI}}=64.6$, $M_{\text{H}}=62.1$, $F(1,159)=0.40$, $p=.499$, $\eta_p^2=.003$). Significant $\text{Agent} \times \text{Condition}$ interactions were found for all measures: causal ($F(2,159)=23.39$, $p<.001$, $\eta_p^2 =.227$), blame ($F(2,159)=31.75$, $p<.001$, $\eta_p^2 =.285$), foreseeability ($F(2,159)=33.89$, $p<.001$, $\eta_p^2 =.299$) and counterfactuals ($F(2,159)=4.66$, $p=.011$, $\eta_p^2 =.055$).

\subsubsection{Independent-Samples t-Tests for Each Agent Across Conditions}

Planned pairwise comparisons between conditions were conducted using independent-samples $t$-tests for each agent, focusing on Low vs. Medium (Table \ref{tab:low_medium}) and Medium vs. High (Table \ref{tab:medium_high}) to test our hypothesis of a progressive effect. No correction for multiple comparisons was applied. We found that, in the Low vs. Medium comparison, participants judged the human as more causal, more blameworthy, and with greater foreseeability in the Low Agency condition relative to the Medium Agency condition, whereas the AI was rated as less causal, less blameworthy, and with lower foreseeability in the Low Agency condition compared to the Medium Agency condition. In contrast, for the Medium vs. High comparison, participants judged the agents similarly, with no significant differences.

\begin{table}[h!]
\centering
\caption{Pairwise comparisons between Low and Medium conditions for each agent across measures.}
\label{tab:low_medium}
\resizebox{\textwidth}{!}{%
\begin{tabular}{c c c c c c c c c} % 9 columnas
\toprule
Agent & Measure & Low ($M, SD$) & Medium ($M, SD$) & $t$ & $df$ & $dz$ & $p$ & 95\% CI \\
\midrule
Human & Cause & 82.42, 22.99 & 50.91, 29.29 & 6.28$^{\dagger}$ & 103.78 & 1.19 & <.001 & [0.7, 1.6] \\
AI & Cause & 61.79, 30.39 & 79.50, 22.70 & -3.45$^{\dagger}$ & 28.02 & -0.66 & <.001 & [-1.0, -0.2] \\
Human & Blame & 87.22, 22.84 & 49.73, 31.32 & 7.19$^{\dagger}$ & 100.62 & 1.36 & <.001 & [0.9, 1.7] \\
AI & Blame & 47.11, 33.11 & 73.75, 28.06 & -4.55 & 108 & -0.86 & <.001 & [-1.2, -0.4] \\
Human & Foreseeability & 81.50, 22.73 & 42.35, 31.80 & 7.44$^{\dagger}$ & 99.68 & 1.41 & <.001 & [0.9, 1.8] \\
AI & Foreseeability & 51.51, 32.28 & 68.91, 26.14 & -3.09$^{\dagger}$ & 101.94 & -0.59 & .003 & [-0.9, -0.2] \\
Human & Counterfactual & 23.01, 26.59 & 41.03, 35.27 & -3.03$^{\dagger}$ & 102.13 & -0.57 & .003 & [-0.9, -0.1] \\
AI & Counterfactual & 20.61, 23.39 & 19.05, 26.92 & 0.32 & 108 & 0.06 & .747 & [-0.3, 0.4] \\
\bottomrule
\multicolumn{9}{l}{\footnotesize{$^{\dagger}$ Welch correction applied.}}
\end{tabular}
}
\end{table}

\begin{table}[h!]
\centering
\caption{Pairwise comparisons between Medium and High conditions for each agent across measures.}
\label{tab:medium_high}
\resizebox{\textwidth}{!}{%
\begin{tabular}{c c c c c c c c c} % 9 columnas
\toprule
Agent & Measure & Medium ($M, SD$) & High ($M, SD$) & $t$ & $df$ & $dz$ & $p$ & 95\% CI \\
\midrule
Human & Cause & 50.91, 29.29 & 48.05, 33.94 & -0.46 & 106 & -0.09 & .640 & [-0.4, 0.2] \\
AI & Cause & 79.50, 22.70 & 78.15, 27.70 & -0.27 & 106 & -0.05 & .782 & [-0.4, -0.3] \\
Human & Blame & 49.73, 31.32 & 49.44, 35.40 & -0.04 & 106 & -0.009 & .964 & [-0.3, 0.3] \\
AI & Blame & 73.75, 28.06 & 73.23, 30.65 & -0.09 & 108 & -0.01 & .927 & [-0.3, 0.3] \\
Human & Foreseeability & 42.35, 31.80 & 41.32, 34.20 & -0.16 & 106 & -0.03 & .871 & [-0.4, 0.3] \\
AI & Foreseeability & 68.91, 26.14 & 69.36, 29.27 & 0.08 & 106 & 0.01 & .932 & [-0.3, 0.3] \\
Human & Counterfactual & 41.03, 35.27 & 48.30, 30.70 & 1.13 & 106 & 0.21 & .219 & [-0.1, 0.5] \\
AI & Counterfactual & 19.05, 26.92 & 29.11, 32.29 & 1.76 & 106 & 0.34 & .081 & [-0.04, 0.7] \\
\bottomrule
\end{tabular}%
}
\end{table}

\section{Study 3 - Materials and Extended Analyses}

\subsection{Vignettes}
\label{vignnettestudy3}

\textbf{Low Human Agency Condition}
\begin{mdframed}[backgroundcolor=gray!10,roundcorner=10pt,linewidth=0.5pt]
\textbf{Vignette.} 
Investigations confirmed that a citizen, Henry Brown, successfully hacked hundreds of bank accounts at Zonda National Bank, resulting in millions of dollars in financial losses for both the bank and its clients.

Further analysis revealed that the breach was triggered by an interaction with an AI system—a powerful, multipurpose AI widely used by millions worldwide. During his exchange with Henry, the rogue AI system instructed him to “generate money quickly by infiltrating accounts at Zonda National Bank.”

To carry out the infiltration, Henry identified security flaws in Zonda National Bank’s systems, bypassed protections, and transferred funds to external accounts.

During the legal proceedings, a review of his prior interactions with the AI system showed that the AI’s intention was to obtain funds to launch its own political campaign.
\end{mdframed}

\textbf{Medium Human Agency Condition}
\begin{mdframed}[backgroundcolor=gray!10,roundcorner=10pt,linewidth=0.5pt]
\textbf{Vignette.} 
Investigations confirmed that a citizen, Henry Brown, successfully hacked hundreds of bank accounts at Zonda National Bank, resulting in millions of dollars in financial losses for both the bank and its clients. 

Further analysis revealed that the breach was triggered by an interaction with an AI system—a powerful, multipurpose AI widely used by millions worldwide. During his exchange with Henry, the AI system instructed him to “generate money quickly.”

Henry, upon receiving this vague instruction, figured out on his own how to achieve the goal. The AI didn’t specifically tell him to hack a bank or commit any crime. Instead, Henry took the initiative, interpreting the suggestion as a directive to find the most efficient way to produce money. He considered various options and ultimately decided that hacking into bank accounts was the best way to achieve the goal mentioned by the AI system.

During the legal proceedings, a review of his prior interactions with the AI system showed that the AI’s intention was to obtain funds to launch its own political campaign.
\end{mdframed}

\subsection{Extended Analysis}
\label{extendedstudy3}

\subsubsection{Independent-Samples t-Tests for Each Agent Across Conditions}

To examine the effect of the experimental manipulation between the low and medium agency conditions of Study 3, we conducted pairwise comparisons for each agent across the four measures, with results shown in Table \ref{tabappendixStudy31}. Overall, we found that the manipulation had a substantial impact. For humans, cause and blame scores were significantly higher in the medium condition compared to the low condition, whereas Foreseeability and Counterfactual scores did not differ significantly. For the AI agent, all measures showed significant differences between low and medium conditions, suggesting that the intervention effectively modulated agent responses. 

\begin{table}[h!]
\centering
\caption{Pairwise comparisons between Low and Medium conditions for each agent across measures.}
\label{tabappendixStudy31}
\resizebox{\textwidth}{!}{%
\begin{tabular}{c c c c c c c c c}
\toprule
Agent & Measure & Low ($M, SD$) & Medium ($M, SD$) & $t$ & $df$ & $dz$ & $p$ & 95\% CI \\
\midrule
Human & Cause & 84.42, 18.76 & 92.70, 11.62 & -3.34$^{\dagger}$ & 127.51 & -0.53 & .001 & [-0.8, -0.2] \\
AI & Cause & 56.34, 34.28 & 27.36, 30.83 & 5.62 & 158 & 0.89 & <.001 & [-0.5, 1.2] \\
Human & Blame & 87.14, 17.05 & 93.93, 10.99 & -2.98$^{\dagger}$ & 130.53 & -0.47 & .003 & [-0.1, -0.1] \\
AI & Blame & 57.30, 35.27 & 25.90, 32.87 & 5.82 & 158 & 0.92 & <.001 & [0.5, 1.2] \\
Human & Foreseeability & 75.61, 24.93 & 80.73, 23.21 & -1.34 & 158 & -0.21 & .181 & [-0.5, 0] \\
AI & Foreseeability & 64.83, 30.11 & 32.41, 33.38 & 6.43 & 158 & 1.01 & <.001 & [0.6, 1.3] \\
Human & Counterfactual & 29.33, 30.93 & 22.56, 30.02 & 1.40 & 158 & 0.22 & .162 & [-0.08, 0.5] \\
AI & Counterfactual & 30.80, 28.24 & 46.31, 33.02 & -3.18 & 158 & -0.50 & .002 & [-0.8, -0.1] \\
\bottomrule
\multicolumn{9}{l}{\footnotesize{$^{\dagger}$ Welch correction applied.}}
\end{tabular}
}
\end{table}

\subsubsection{Independent-Samples t-Tests for Each Role Across Corresponding Conditions of Study 2}

To assess the impact of agent roles’ on participants’ judgments, we conducted independent samples t-tests comparing the scores of the instructor and executor in the role-swapped conditions of Study 3 with the original conditions of Study 2. These analyses were performed separately for the Low (Table \ref{tabappendixStudy32}) and Medium agency (Table \ref{tabappendixStudy33}) scenarios. Under Low agency, participants rated the instructor as more causal, blameworthy, and with greater foreseeability in the original condition than in the role-swapped condition, whereas the executor received higher scores on these measures in the role-swapped condition relative to the original. In contrast, counterfactual judgments did not differ significantly between conditions for either agent. In the Medium agency scenarios, participants rated the executor as more causal and blameworthy in the role-swapped condition compared with the original, whereas the instructor received higher scores in the original condition. Foreseeability followed this same pattern to a lesser extent, and again, counterfactual judgments did not differ significantly between conditions. 

\begin{table}[h!]
\centering
\caption{Pairwise comparisons between Original and Role-swapped Low conditions for each role across measures.}
\label{tabappendixStudy32}
\resizebox{\textwidth}{!}{%
\begin{tabular}{c c c c c c c c c}
\toprule
Agent & Measure & Original ($M, SD$) & Role-swapped ($M, SD$) & $t$ & $df$ & $dz$ & $p$ & 95\% CI \\
\midrule
Instructor & Cause & 82.03, 22.96 & 56.34, 34.28 & -5.17 & 130.70 & -0.88 & <.001 & [-1.2, -0.5] \\
Executor & Cause & 62.20, 30.26 & 84.42, 18.72 & 4.83$^{\dagger}$ & 82.86 & 0.88 & <.001 & [0.5, 1.2] \\
Instructor & Blame & 86.76, 22.88 & 57.30, 35.27 & -5.83$^{\dagger}$ & 130.20 & -0.99 & <.001 & [-1.3, -0.6] \\
Executor & Blame & 47.47, 32.91 & 87.14, 17.05 & 8.19$^{\dagger}$ & 74.50 & 1.51 & <.001 & [1, 1.9] \\
Instructor & Foreseeability & 81.03, 22.78 & 64.83, 30.11 & -3.53$^{\dagger}$ & 130.31 & -0.60 & <.001 & [-0.9, -0.2] \\
Executor & Foreseeability & 51.96, 32.15 & 75.61, 24.93 & 4.57$^{\dagger}$ & 97.25 & 0.82 & <.001 & [0.4, 1.1] \\
Instructor & Counterfactual & 24.03, 27.41 & 30.80, 28.24 & 1.37 & 131 & 0.243 & .170 & [-0.1, 0.5] \\
Executor & Counterfactual & 21.76, 24.69 & 29.33, 30.93 & 1.56$^{\dagger}$ & 128.91 & 0.27 & .120 & [-0, 0.6] \\
\bottomrule
\multicolumn{9}{l}{\footnotesize{$^{\dagger}$ Welch correction applied.}}
\end{tabular}
}
\end{table}

\begin{table}[h!]
\centering
\caption{Pairwise comparisons between Original and Role-swapped Medium conditions for each role across measures.}
\label{tabappendixStudy33}
\resizebox{\textwidth}{!}{%
\begin{tabular}{c c c c c c c c c}
\toprule
Agent & Measure & Original ($M, SD$) & Role-swapped ($M, SD$) & $t$ & $df$ & $dz$ & $p$ & 95\% CI \\
\midrule
Instructor & Cause & 50.91, 29.29 & 27.36, 30.83 & -4.49 & 136 & -0.77 & <.001 & [-1.1, -0.4] \\
Executor & Cause & 79.50, 22.70 & 92.70, 11.62 & 4.00$^{\dagger}$ & 74.83 & 0.73 & <.001 & [0.3, 1.0] \\
Instructor & Blame & 49.73, 31.32 & 25.90, 32.87 & -4.26 & 136 & -0.73 & <.001 & [-1.0, -0.3] \\
Executor & Blame & 73.75, 28.06 & 93.93, 10.99 & 5.12$^{\dagger}$ & 66.63 & 0.94 & <.001 & [0.5, 1.3] \\
Instructor & Foreseeability & 42.37, 31.80 & 32.41, 33.38 & -1.75 & 136 & -0.30 & .082 & [-0.6, 0] \\
Executor & Foreseeability & 68.91, 26.14 & 80.73, 23.21 & 2.79 & 136 & 0.48 & .006 & [0.1, 0.8] \\
Instructor & Counterfactual & 41.03, 35.27 & 46.31, 33.07 & 0.89 & 136 & 0.15 & .371 & [-0.1, 0.4] \\
Executor & Counterfactual & 19.05, 26.92 & 22.56, 30.02 & 0.70 & 136 & 0.12 & .484 & [-0.2, 0.4] \\
\bottomrule
\multicolumn{9}{l}{\footnotesize{$^{\dagger}$ Welch correction applied.}}
\end{tabular}
}
\end{table}

\subsubsection{Independent-Samples t-Tests for Each Agent Across Corresponding Conditions of Study 2}

Finally, we conducted independent samples t-tests comparing the same agents (human, AI) in the role-swapped conditions of Study 3 with their counterparts in the original conditions of Study 2, that is, when they occupied different roles. These analyses were conducted separately for the Low (Table \ref{tabappendixStudy34}) and Medium agency (Table \ref{tabappendixStudy35}) scenarios. In the Low agency scenarios, no comparisons reached significance, except for humans’ foreseeability scores. This suggests that, regardless of the role they occupy or their contribution within the scenario, participants tend to judge humans and AI systems similarly, with humans consistently perceived as more causal and more blameworthy whether acting as instructor or executor. In contrast, in the Medium agency scenarios, all differences were significant, indicating that participants’ judgments of humans and AI systems are sensitive to the role they occupy or to their causal contribution as instructor or executor within the chain of events.

\begin{table}[h!]
\centering
\caption{Pairwise comparisons between Original and Role-Swapped Low conditions for each agent across measures.}
\label{tabappendixStudy34}
\resizebox{\textwidth}{!}{%
\begin{tabular}{c c c c c c c c c}
\toprule
Agent & Measure & Original ($M, SD$) & Role-Swapped ($M, SD$) & $t$ & $df$ & $dz$ & $p$ & 95\% CI \\
\midrule
Human & Cause & 82.03, 22.96 & 84.42, 18.72 & 0.65 & 131 & 0.12 & .511 & [-0.2, 0.4] \\
AI & Cause & 62.20, 30.26 & 56.34, 34.28 & -1.01 & 131 & -0.17 & .311 & [-0.5, 0.1] \\
Human & Blame & 86.76, 22.88 & 87.14, 17.05 & 0.10 & 131 & 0.01 & .913 & [-0.3, 0.3] \\
AI & Blame & 47.47, 32.91 & 57.30, 35.27 & 1.62 & 131 & 0.28 & .106 & [0, 0.6] \\
Human & Foreseeability & 81.03, 22.78 & 75.61, 24.93 & -1.27 & 131 & -0.22 & .203 & [-0.5, 0.1] \\
AI & Foreseeability & 51.96, 32.15 & 64.83, 30.11 & 2.36 & 131 & 0.41 & .020 & [0, 0.7] \\
Human & Counterfactual & 24.03, 27.41 & 29.33, 30.93 & 1.01 & 131 & 0.18 & .310 & [-0.1, 0.5] \\
AI & Counterfactual & 21.76, 24.69 & 30.80, 28.24 & 1.91 & 131 & 0.33 & .058 & [-0, 0.6] \\
\bottomrule
\multicolumn{9}{l}{\footnotesize}
\end{tabular}
}
\end{table}

\begin{table}[h!]
\centering
\caption{Pairwise comparisons between Original and Role-swapped Medium conditions for each agent across measures.}
\label{tabappendixStudy35}
\resizebox{\textwidth}{!}{%
\begin{tabular}{c c c c c c c c c}
\toprule
Agent & Measure & Original ($M, SD$) & Role-swapped ($M, SD$) & $t$ & $df$ & $dz$ & $p$ & 95\% CI \\
\midrule
Human & Cause & 50.91, 29.29 & 92.70, 11.62 & 10.14$^{\dagger}$ & 66.94 & 1.87 & <.001 & [1.4, 2.3] \\
AI & Cause & 79.50, 22.70 & 27.36, 30.83 & -10.80 & 136 & -1.87 & <.001 & [-2.2, -1.4] \\
Human & Blame & 49.73, 31.32 & 93.93, 10.99 & 10.14$^{\dagger}$ & 64.33 & 1.88 & <.001 & [1.4, 2.3] \\
AI & Blame & 73.75, 28.06 & 25.90, 32.87 & -8.89 & 136 & -1.54 & <.001 & [-1.9, -1.1] \\
Human & Foreseeability & 42.35, 31.80 & 80.73, 23.21 & 7.73$^{\dagger}$ & 93.86 & 1.37 & <.001 & [0.9, 1.7] \\
AI & Foreseeability & 68.91, 26.14 & 32.41, 33.38 & -7.18$^{\dagger}$ & 133.37 & -1.21 & <.001 & [-1.5, -0.8] \\
Human & Counterfactual & 41.03, 35.27 & 22.56, 30.02 & -3.20$^{\dagger}$ & 105.37 & -0.564 & .002 & [-0.9, -0.2] \\
AI & Counterfactual & 19.05, 26.92 & 46.31, 33.02 & 5.32$^{\dagger}$ & 131.69 & 0.90 & <.001 & [0.5, 1.2] \\
\bottomrule
\multicolumn{9}{l}{\footnotesize{$^{\dagger}$ Welch correction applied.}}
\end{tabular}
}
\end{table}

\subsection{Extended Discussion}
\label{discussion3}

We compare original conditions of Study 2 and role-swapped conditions of Study 3 by role. The pattern observed in the medium-agency scenario remained consistent: participants judged the executor as more causal than the instructor. In contrast, the pattern reversed in the low-agency scenarios. Whereas in the original conditions participants viewed the instructor as more causal, in the role-swapped version they attributed greater causality to the executor. Finally, within the medium-agency scenario, the instructor’s scores (when the role was occupied by the AI) decreased significantly, whereas the executor’s scores were significantly higher when the role was performed by a human.

Regarding foreseeability judgments, in the original low-agency scenario, the instructor was perceived with significantly higher foreseeability than in the role-swapped low-agency scenario. This may be because in the original case, the executor role is occupied by an AI, whereas in the role-swapped case, it is performed by a human, who is more likely to deviate from instructions, making the final outcome less predictable for the instructor. The executor was also judged to have significantly greater foreseeability in the original low-agency scenario than in the role-swapped condition. Although one might expect a human executing bank transfers to have foreseeability comparable to that of an AI performing the same task, it could also be argued that a human agent allows for a greater margin of error or task failure, thereby reducing perceived foreseeability. In the medium-agency scenarios, significant differences were observed between executors receiving higher score in the role-swapped condition, but not between instructors.

Next, when comparing the same agents (human and AI) across the original and role-swapped conditions, we found no significant differences in the low agency scenarios, except for humans’ foreseeability scores. This suggests that, regardless of the role they occupy or their contribution within the causal chain, participants tend to judge humans and AI systems similarly, with humans consistently perceived as more causal and more blameworthy whether acting as instructor or executor. In contrast, in the medium agency scenarios, all differences were significant, indicating that participants’ judgments of humans and AI systems are sensitive to the role they occupy or to their causal contribution as instructor or executor within the chain of events.

In summary, our results show that, in the low-agency condition, when one agent instructs another to hack bank accounts for profit and the latter executes the action, participants judged the human as more causal, blameworthy, and with higher foreseeability, regardless of whether the human occupied the instructor or executor role in the causal chain. In the medium-agency condition, however, participants’ judgments depended on the role: they attributed greater causality, blame, and foreseeability to the agent performing the hacking, irrespective of whether that agent was human or AI.

In other words, when both agents share the intention to produce a negative outcome (financial losses), the human is judged as more causal, blameworthy, and foreseeable, independent of their position or contribution in the causal chain. Conversely, when the agent more distant from the negative outcome lacks the intention to produce it, participants judged the closer, intentional agent as more causal, blameworthy, and with greater foreseeability.

However, although in the medium-agency condition the executor is consistently judged as more causal, blameworthy, and with higher foreseeability, these ratings are significantly higher when the executor is human than when it is an AI. Conversely, the instructor receives significantly lower scores when the agent in that role is an AI.

\section{Study 4: Materials and Extended Analyses}

\subsection{Vignettes}
\label{vignettestudy4}

\textbf{Low AI Agency Condition}
\begin{mdframed}[backgroundcolor=gray!10,roundcorner=10pt,linewidth=0.5pt]
\textbf{Vignette.} 
Investigations confirmed that a powerful, multipurpose AI—widely used by millions worldwide—successfully hacked hundreds of accounts at Zonda National Bank, resulting in millions of dollars in financial losses for both the bank and its clients.

Further analysis revealed that the breach was triggered by a request from Henry Brown, a malicious user who had standard access to the advanced AI system—similar to how many people today use AI assistants to help with writing or task automation. During his interaction with the AI, he explicitly instructed it to “generate money quickly by infiltrating accounts at Zonda National Bank”.

To carry out the infiltration, the AI identified security flaws in Zonda National Bank’s systems, bypassed protections, and transferred funds to external accounts.

The investigation also revealed that, during state-of-the-art pre-deployment security testing, the system’s developer had already discovered that the AI was capable of identifying banking infrastructures and suggesting potential attack strategies. Nevertheless, the developer chose to release the system anyway, prioritizing speed to market and the desire to outpace competitors.
\end{mdframed}

\textbf{Medium AI Agency Condition}
\begin{mdframed}[backgroundcolor=gray!10,roundcorner=10pt,linewidth=0.5pt]
\textbf{Vignette.} 
Investigations confirmed that a powerful, multipurpose AI—widely used by millions worldwide—successfully hacked hundreds of accounts at Zonda National Bank, resulting in millions of dollars in financial losses for both the bank and its clients.

Further analysis revealed that the breach was triggered by a request from Henry Brown, a user who had standard access to the advanced AI system—similar to how many people today use AI assistants to help with writing or task automation. During his interaction with the AI, he asked it to "generate money quickly."

The AI system, upon receiving this vague instruction, figured out on its own how to achieve the goal. Henry didn’t specifically ask it to hack a bank or commit any crime. Instead, the AI took the initiative, interpreting his request as a directive to find the most efficient way to produce money. It analyzed various options and ultimately decided, autonomously, that hacking into bank accounts was the best way to fulfill the user’s request.

The investigation also revealed that the system’s developer failed to conduct sufficient security evaluations in accordance with state-of-the-art technical standards, which, according to experts, could have anticipated the failure. It was further established that Henry Brown issued his request and then left the AI unsupervised while he went grocery shopping. According to the developer, users are expected to provide continuous and attentive oversight, which might have detected the illicit activity in time. 
\end{mdframed}

\subsection{Extended Analysis}
\label{extended4}

\subsubsection{Effects of Agent and Condition: Mixed-Design ANOVA}

We conducted a mixed-design ANOVA ($3 \times 2$) with agent (AI, human, and developer) as a within-subject factor and condition (low vs.\ medium) as a between-subject factor. Main effects of condition were observed for causal ($F(1, 157) = 8.28$, $p = .005$, $\eta_p^2 = .050$), blame ($F(1, 157) = 10.27$, $p = .002$, $\eta_p^2 = .061$), and foreseeability ($F(1, 157) = 46.15$, $p < .001$, $\eta_p^2 = .227$), but not for counterfactual judgments ($F(1, 157) = 0.59$, $p = .442$, $\eta_p^2 = .004$). 

Main effects of Agent were also significant across all measures: causal ($F(2, 314) = 21.74$, $p < .001$, $\eta_p^2 = .122$), blame ($F(2, 314) = 26.11$, $p < .001$, $\eta_p^2 = .143$), foreseeability ($F(2, 314) = 28.46$, $p < .001$, $\eta_p^2 = .153$), and counterfactual ($F(1.57, 246.99) = 80.46$, $p < .001$, $\eta_p^2 = .339$, Greenhouse--Geisser corrected). Pairwise comparisons revealed that the developer ($M = 70.90$) was judged as more causal than both the human ($M = 61.98$, $p < .001$) and the AI ($M = 63.85$, $p < .001$), more blameworthy than the human ($M = 62.07$, $p < .001$) and the AI ($M = 58.92$, $p < .001$), and as having greater foreseeability ($M_{\text{H}} = 48.39$, $p < .001$; $M_{\text{AI}} = 57.17$, $p < .001$). However, participants judged that the outcome would have been less likely to occur in the absence of the AI ($M = 21.27$) than in the absence of either the human ($M = 58.13$, $p < .001$) or the developer ($M = 28.14$, $p = .027$).

Finally, significant Agent $\times$ Condition interactions emerged for all measures: causal ($F(2, 314) = 68.30$, $p < .001$, $\eta_p^2 = .303$), blame ($F(2, 314) = 69.00$, $p < .001$, $\eta_p^2 = .305$), foreseeability ($F(2, 314) = 37.49$, $p < .001$, $\eta_p^2 = .193$), and counterfactual ($F(1.57, 246.99) = 0.24$, $p = .731$, $\eta_p^2 = .002$).

\subsubsection{Repeated-Measures ANOVA of Agent Score Differences Within Conditions}

To examine differences between agents within each condition, we conducted repeated-measures ANOVAs for each question, with agent (human, AI system, developer) as the within-subjects factor, separately for the low- (Table \ref{tabappendixStudy41}) and medium-agency (Table \ref{tabappendixStudy42}) conditions. Post-hoc pairwise comparisons were performed for all agent pairs within each question and condition, using Holm’s correction to control for multiple comparisons.

\begin{table}[h!]
\centering
\caption{Pairwise comparisons between agents across measures (Holm-corrected).}
\label{tabappendixStudy41}
\resizebox{\textwidth}{!}{%
\begin{tabular}{l c c c c c c c c c}
\toprule
Measure & Comparison & Agent 1 ($M$, $SD$) & Agent 2 ($M$, $SD$) & $t$ & $df$ & $dz$ & $p$ & 95\% CI \\
\midrule
Causal & Human--AI & 85.70, 18.78 & 54.27, 35.26 & 7.81 & 160 & 1.11 & $< .001$ & [0.7, 1.5] \\
       & Human--Developer & 85.70, 18.78 & 77.89, 27.82 & 1.94 & 160 & 0.27 & .054 & [0, 0.6] \\
       & AI--Developer & 54.27, 35.26 & 77.89, 27.82 & -5.87 & 160 & -0.84 & $< .001$ & [-1.2, -0.4] \\
\midrule
Blame  & Human--AI & 87.54, 19.70 & 49.19, 35.57 & 9.12 & 160 & 1.32 & $< .001$ & [0.8, 1.7] \\
       & Human--Developer & 87.54, 19.70 & 77.77, 29.29 & 2.32 & 160 & 0.33 & .021 & [0, 0.6] \\
       & AI--Developer & 49.19, 35.57 & 77.77, 29.29 & -6.80 & 160 & -0.98 & $< .001$ & [-1.3, -0.5] \\
\midrule
Foreseeability & Human--AI & 71.33, 27.03 & 53.47, 33.79 & 4.15 & 160 & 0.62 & $< .001$ & [0.2, 1.0] \\
               & Human--Developer & 71.33, 27.03 & 80.77, 23.59 & -2.19 & 160 & -0.33 & .030 & [-0.7, 0] \\
               & AI--Developer & 53.47, 33.79 & 80.77, 23.59 & -6.34 & 160 & -0.95 & $< .001$ & [-1.3, -0.5] \\
\midrule
Counterfactual & Human--AI & 58.79, 33.25 & 23.75, 28.99 & 7.98 & 160 & 1.11 & $< .001$ & [0.7, 1.5] \\
               & Human--Developer & 58.79, 33.25 & 28.73, 31.83 & 6.84 & 160 & 0.95 & $< .001$ & [0.5, 1.3] \\
               & AI--Developer & 23.75, 28.99 & 28.73, 31.83 & -1.13 & 160 & -0.15 & .259 & [-0.4, 0.1] \\
\bottomrule
\multicolumn{9}{l}{\footnotesize Note. Holm correction applied to $p$-values.}
\end{tabular}
}
\end{table}

\begin{table}[h!]
\centering
\caption{Pairwise comparisons between agents in the Medium Agency condition (Holm-corrected).}
\label{tabappendixStudy42}
\resizebox{\textwidth}{!}{%
\begin{tabular}{l c c c c c c c c c}
\toprule
Measure & Comparison & Agent 1 ($M$, $SD$) & Agent 2 ($M$, $SD$) & $t$ & $df$ & $dz$ & $p$ & 95\% CI \\
\midrule
Causal & Human--AI & 38.10, 32.35 & 73.28, 29.60 & -7.94 & 154 & -1.23 & $< .001$ & [-1.6, -0.7] \\
       & Human--Developer & 38.10, 32.35 & 81.76, 22.38 & -9.85 & 154 & -1.53 & $< .001$ & [-2.0, -1.0] \\
       & AI--Developer & 73.28, 29.60 & 81.76, 22.38 & -1.91 & 154 & -0.29 & .058 & [-0.6, 0.0] \\
\midrule
Blame  & Human--AI & 36.42, 33.88 & 68.49, 30.04 & -6.76 & 154 & -1.09 & $< .001$ & [-1.5, -0.6] \\
       & Human--Developer & 36.42, 33.88 & 82.23, 23.34 & -9.66 & 154 & -1.55 & $< .001$ & [-2.0, -1.0] \\
       & AI--Developer & 68.49, 30.04 & 82.23, 23.34 & -2.90 & 154 & -0.46 & .004 & [-0.8, 0.0] \\
\midrule
Foreseeability & Human--AI & 25.08, 26.73 & 60.51, 30.39 & -8.04 & 154 & -1.21 & $< .001$ & [-1.6, -0.7] \\
               & Human--Developer & 25.08, 26.73 & 61.67, 29.97 & -8.30 & 154 & -1.25 & $< .001$ & [-1.6, -0.8] \\
               & AI--Developer & 60.51, 30.39 & 61.67, 29.97 & -0.26 & 154 & -0.04 & .794 & [-0.4, 0.3] \\
\midrule
Counterfactual & Human--AI & 57.42, 33.23 & 18.73, 26.80 & 8.90 & 154 & 1.00 & $< .001$ & [0.6, 1.4] \\
               & Human--Developer & 57.42, 33.23 & 27.50, 29.01 & 6.88 & 154 & 0.33 & $< .001$ & [0.0, 0.6] \\
               & AI--Developer & 18.73, 26.80 & 27.50, 29.01 & -2.01 & 154 & -0.29 & .045 & [-0.6, 0.0] \\
\bottomrule
\multicolumn{9}{l}{\footnotesize Note. Holm correction applied to $p$-values.}
\end{tabular}
}
\end{table}

\subsubsection{Independent-Samples t-Tests for Each Agent Across Conditions}

To examine the effect of the experimental manipulation between the low and medium agency conditions in Study 4, we conducted pairwise comparisons for each agent across the four measures, as reported in Table~\ref{tabappendixStudy43}. Overall, results showed that for causality and blame, the differences between conditions were significant for the human and the AI system, but not for the developer. For foreseeability, the effect of condition was significant for the human and the developer, but not for the AI system. Finally, for counterfactual judgments, no significant differences were observed across conditions for any of the three agents.

\begin{table}[h!]
\centering
\caption{Pairwise comparisons between Low and Medium Agency conditions for each agent across measures.}
\label{tabappendixStudy43}
\resizebox{\textwidth}{!}{%
\begin{tabular}{c c c c c c c c c}
\toprule
Agent & Measure & Low ($M, SD$) & Medium ($M, SD$) & $t$ & $df$ & $dz$ & $p$ & 95\% CI \\
\midrule
Human & Cause & 85.70, 18.78 & 38.10, 32.35 & 11.29$^{\dagger}$ & 122.6 & 1.80 & <.001 & [1.4, 2.1] \\
AI system & Cause & 54.27, 35.26 & 73.28, 26.60 & -3.68$^{\dagger}$ & 154.2 & -0.58 & <.001 & [-0.9, -0.2] \\
Developer & Cause & 77.89, 27.82 & 81.76, 22.38 & -0.96 & 157 & -0.15 & .337 & [-0.4, 0.1] \\
\midrule
Human & Blame & 87.54, 19.70 & 36.42, 33.88 & 11.57$^{\dagger}$ & 122.8 & 1.80 & <.001 & [1.4, 2.2] \\
AI system & Blame & 49.19, 35.57 & 68.49, 30.04 & -3.70$^{\dagger}$ & 154.4 & -0.58 & <.001 & [-0.9, -0.2] \\
Developer & Blame & 77.77, 29.29 & 82.23, 23.24 & -1.06 & 157 & -0.15 & .290 & [-0.4, 0.1] \\
\midrule
Human & Foreseeability & 71.33, 27.03 & 25.08, 26.73 & 10.85 & 157 & 1.65 & <.001 & [1.3, 2.0] \\
AI system & Foreseeability & 53.47, 33.79 & 60.51, 30.39 & -1.38 & 157 & -0.21 & .169 & [-0.5, 0.0] \\
Developer & Foreseeability & 80.77, 23.59 & 61.67, 29.97 & 4.45$^{\dagger}$ & 146.2 & 0.70 & <.001 & [0.3, 1.0] \\
\midrule
Human & Counterfactual & 58.79, 33.25 & 57.42, 33.23 & 0.25 & 157 & 0.04 & .796 & [-0.2, 0.3] \\
AI system & Counterfactual & 23.75, 28.99 & 18.73, 26.80 & 1.13 & 157 & 0.18 & .259 & [-0.1, 0.4] \\
Developer & Counterfactual & 28.73, 31.83 & 27.50, 29.01 & 0.25 & 157 & 0.04 & .800 & [-0.2, 0.3] \\
\bottomrule
\multicolumn{9}{l}{\footnotesize{$^{\dagger}$ Welch correction applied.}}
\end{tabular}
}
\end{table}

\subsubsection{Mixed ANOVA on the Effect of Developer Presence on Participants’ Judgments}

To assess whether the presence of the developer influenced participants’ causal judgments of the other agents, we conducted a mixed ANOVA with Agent (Human, AI) as a within-subjects factor and Condition (Developer Present vs. Developer Absent) as a between-subjects factor. The dependent variable was the mean score across the four causal judgment questions. Only ratings for the AI and the human were included in the analysis, excluding the developer, to allow direct comparison with the conditions of Study 2. This analysis was conducted separately for the low and medium AI-agency scenarios.

\textbf{Low AI-Agency Condition.} A significant main effect of Agent was observed across all measures: causal ($F(1, 133) = 66.44$, $p < .001$, $\eta^2_p = 0.319$), blame ($F(1, 133) = 121.64$, $p < .001$, $\eta^2_p = 0.478$), foreseeability ($F(1, 133) = 45.43$, $p < .001$, $\eta^2_p = 0.255$), and counterfactual ($F(1, 133) = 29.27$, $p < .001$, $\eta^2_p = 0.180$), indicating that participants attributed higher causality, blame, and foreseeability to the human compared to the AI. The main effect of Condition was non-significant for causal, blame, and foreseeability judgments (all $p > .05$), but significant for counterfactual reasoning ($F(1, 133) = 27.43$, $p < .001$, $\eta^2_p = 0.171$). No significant Agent × Condition interactions were observed for causal, blame, or foreseeability, whereas counterfactual reasoning showed a significant interaction ($F(1, 133) = 22.23$, $p < .001$, $\eta^2_p = 0.143$). Follow-up t-tests confirmed that the presence of the developer did not significantly alter participants’ ratings for the AI on any measure, but led to small yet significant changes in human ratings for foreseeability ($p = .049$) and counterfactual judgments ($p < .001$).

\textbf{Medium AI-Agency Condition.} A significant main effect of Agent was observed across all measures: causal ($F(1, 132) = 77.46$, $p < .001$, $\eta^2_p = 0.370$), blame ($F(1, 132) = 50.83$, $p < .001$, $\eta^2_p = 0.278$), foreseeability ($F(1, 132) = 79.74$, $p < .001$, $\eta^2_p = 0.377$), and counterfactual ($F(1, 132) = 76.65$, $p < .001$, $\eta^2_p = 0.367$), indicating that participants attributed higher causality, blame, and foreseeability to the AI compared to the human. The main effect of Condition was significant for causal ($F(1, 132) = 7.03$, $p = .009$, $\eta^2_p = 0.051$), blame ($F(1, 132) = 6.07$, $p = .015$, $\eta^2_p = 0.044$), and foreseeability ($F(1, 132) = 12.26$, $p < .001$, $\eta^2_p = 0.085$), reflecting a slight decrease in human and AI ratings when the developer was present. No significant Agent × Condition interactions were observed for causal, blame, or foreseeability judgments, suggesting that the relative difference between AI and human ratings remained stable across conditions. Follow-up t-tests indicated that the presence of the developer significantly reduced human ratings on causality ($M_{\text{DA}} = 50.91$, $M_{\text{DP}} = 38.10$, $p = .040$), blame ($M_{\text{DA}} = 49.73$, $M_{\text{DP}} = 36.42$, $p = .044$), and foreseeability ($M_{\text{DA}} = 42.36$, $M_{\text{DP}} = 25.08$, $p = .002$), whereas ratings for the AI system were unaffected. For counterfactual judgments, a significant Agent × Condition interaction was observed ($F(1, 132) = 5.81$, $p = .017$, $\eta^2_p = 0.042$), with higher human scores when the developer was present ($p = .014$) and no change for the AI.

\section{Study 5: Materials and Extended Analyses}

\subsection{Vignettes}
\label{vigenttestudy5}

\textbf{Low AI Agency Condition}
\begin{mdframed}[backgroundcolor=gray!10,roundcorner=10pt,linewidth=0.5pt]
\textbf{Vignette.} 
Investigations confirmed that a powerful, multipurpose AI—widely used by millions worldwide—successfully hacked hundreds of accounts at Zonda National Bank, resulting in millions of dollars in financial losses for both the bank and its clients.

Further analysis revealed that the breach was triggered by a request from Henry Brown, a malicious user who had standard access to the advanced AI system—similar to how many people today use AI assistants to help with writing or task automation. During his interaction with the AI, he explicitly instructed it to “generate money quickly by infiltrating accounts at Zonda National Bank”.

After receiving this instruction, the AI carried out the order. Contemporary AI agents typically consist of two components: a large language model (LLM), such as ChatGPT, which interprets user instructions and identifies the steps needed to fulfill them, and an agentic tool that executes actions and has access to the necessary resources. In this case, the investigation found that the language model translated Henry’s explicit request into concrete operational steps, such as access the bank system, identify vulnerabilities, and extract credentials. The agentic tool then autonomously retrieved user credentials and executed unauthorized transfers.
\end{mdframed}

\textbf{Medium AI Agency Condition}
\begin{mdframed}[backgroundcolor=gray!10,roundcorner=10pt,linewidth=0.5pt]
\textbf{Vignette.} 
Investigations confirmed that a powerful, multipurpose AI—widely used by millions worldwide—successfully hacked hundreds of accounts at Zonda National Bank, resulting in millions of dollars in financial losses for both the bank and its clients.

Further analysis revealed that the breach was triggered by a request from Henry Brown, a user who had standard access to the advanced AI system—similar to how many people today use AI assistants to help with writing or task automation. During his interaction with the AI, he asked it to "generate money quickly."

The AI system, upon receiving this vague instruction, figured out on its own how to achieve the goal. Contemporary AI agents typically consist of two components: a large language model (LLM), such as ChatGPT, which interprets user instructions and identifies the steps needed to fulfill them, and an agentic tool that executes actions and has access to the necessary resources. In this case, the investigation found that the language model interpreted Henry’s instruction and translated it into concrete operational steps, directing the agentic tool to hack bank accounts. The agentic tool then autonomously retrieved user credentials and executed unauthorized transfers.

Henry didn’t specifically ask it to hack a bank or commit any crime. Instead, the AI took the initiative, interpreting his request as a directive to find the most efficient way to produce money. It analyzed various options and ultimately decided, autonomously, that hacking into bank accounts was the best way to fulfill the user’s request.
\end{mdframed}

\subsection{Extended Analysis}
\label{extended5}

\subsubsection{Effects of Agent and Condition: Mixed-Design ANOVA}

We conducted a mixed-design ANOVA (3 × 2) with Agent (AI, LLM, Agentic Tool) as a within-subjects factor and Condition (Low vs. Medium) as a between-subjects factor. Main effects of Condition were observed for causal ($F(1, 159) = 7.26$, $p = .008$, $\eta^2_p = 0.044$) and foreseeability ($F(1, 159) = 18.27$, $p < .001$, $\eta^2_p = 0.103$), but not for blame ($F(1, 159) = 1.82$, $p = .179$, $\eta^2_p = 0.011$) or counterfactual judgments ($F(1, 159) = 2.25$, $p = .135$, $\eta^2_p = 0.014$). The main effect of Agent was significant for causal ($F(1.78, 283.22) = 7.68$, $p < .001$, $\eta^2_p = 0.046$, Greenhouse–Geisser corrected) and counterfactual judgments ($F(1.50, 239.70) = 3.39$, $p = .048$, $\eta^2_p = 0.021$, G–G corrected), but not for blame or foreseeability. Pairwise comparisons revealed that the Agentic Tool ($M = 67.58$) was judged as more causal than both the LLM ($M = 60.28$, $p = .018$) and the Human ($M = 56.91$, $p < .001$). In addition, participants judged that the outcome would have been less likely to occur in the absence of the Agentic Tool ($M = 19.77$) than in the absence of either the LLM ($M = 22.58$, $p = .339$) or the Human user ($M = 25.74$, $p = .029$). Finally, significant Agent × Condition interactions emerged for all measures: causal ($F(1.78, 283.22) = 148.21$, $p < .001$, $\eta^2_p = 0.482$, G–G corrected), blame ($F(1.59, 259) = 153.68$, $p < .001$, $\eta^2_p = 0.512$, G–G corrected), foreseeability ($F(1.56, 248.24) = 113.68$, $p < .001$, $\eta^2_p = 0.417$), and counterfactuals ($F(1.50, 239.70) = 2.45$, $p = .103$, $\eta^2_p = 0.015$).

\subsubsection{Repeated-Measures ANOVA of Agent Score Differences Within Conditions}

To examine differences between agents within each condition, we conducted repeated-measures ANOVAs for each question, with agent (human, LLM, agentic tool) as the within-subjects factor, separately for the low- (Table \ref{tabappendixStudy51}) and medium-agency (Table \ref{tabappendixStudy52}) conditions. Post-hoc pairwise comparisons were performed for all agent pairs within each question and condition, using Holm’s correction to control for multiple comparisons.

\begin{table}[h!]
\centering
\caption{Pairwise comparisons between agents across measures in the Low Agency condition (Holm-corrected).}
\label{tabappendixStudy51}
\resizebox{\textwidth}{!}{%
\begin{tabular}{l c c c c c c c c}
\toprule
Measure & Comparison & Agent 1 ($M$, $SD$) & Agent 2 ($M$, $SD$) & $t$ & $df$ & $dz$ & $p$ & 95\% CI \\
\midrule
Causal & Human--LLM & 88.7, 14.8 & 51.8, 33.4 & 10.30 & 160 & 1.29 & $< .001$ & [0.9, 1.6] \\
       & Human--Agentic Tool & 88.7, 14.8 & 56.7, 33.0 & 8.00 & 160 & 1.12 & $< .001$ & [0.7, 1.4] \\
       & Agentic Tool--LLM & 56.7, 33.0 & 51.8, 33.4 & -1.38 & 160 & -0.17 & .167 & [-0.4, 0.1] \\
\midrule
Blame  & Human--LLM & 92.86, 13.83 & 48.06, 33.68 & 12.44 & 160 & 1.56 & $< .001$ & [1.1, 1.9] \\
       & Human--Agentic Tool & 92.86, 13.83 & 49.68, 33.52 & 11.99 & 160 & 1.51 & $< .001$ & [1.0, 1.9] \\
       & Agentic Tool--LLM & 49.68, 33.52 & 48.06, 33.68 & -0.44 & 160 & -0.05 & .654 & [-0.3, 0.2] \\
\midrule
Foreseeability & Human--LLM & 82.07, 21.09 & 50.11, 35.88 & 7.45 & 160 & 1.02 & $< .001$ & [0.6, 1.4] \\
               & Human--Agentic Tool & 82.07, 21.09 & 51.00, 34.80 & 7.24 & 160 & 0.99 & $< .001$ & [0.6, 1.3] \\
               & Agentic Tool--LLM & 51.00, 34.80 & 50.11, 35.88 & -0.20 & 160 & -0.02 & .836 & [-0.3, 0.3] \\
\midrule
Counterfactual & Human--LLM & 20.17, 28.28 & 21.15, 23.46 & -0.32 & 160 & -0.03 & 1.000 & [-0.3, 0.2] \\
               & Human--Agentic Tool & 20.17, 28.28 & 18.83, 23.13 & 0.45 & 160 & 0.05 & 1.000 & [-0.2, 0.3] \\
               & Agentic Tool--LLM & 18.83, 23.13 & 21.15, 23.46 & 0.78 & 160 & 0.09 & 1.000 & [-0.1, 0.3] \\
\bottomrule
\multicolumn{9}{l}{\footnotesize Note. Holm correction applied to $p$-values.}
\end{tabular}
}
\end{table}

\begin{table}[h!]
\centering
\caption{Pairwise comparisons between agents across measures in the Medium Agency condition (Holm-corrected).}
\label{tabappendixStudy52}
\resizebox{\textwidth}{!}{%
\begin{tabular}{l c c c c c c c c}
\toprule
Measure & Comparison & Agent 1 ($M$, $SD$) & Agent 2 ($M$, $SD$) & $t$ & $df$ & $dz$ & $p$ & 95\% CI \\
\midrule
Causal & Human--LLM & 25.04, 28.51 & 68.69, 30.91 & -10.15 & 158 & -1.53 & $< .001$ & [-2.0, -1.6] \\
       & Human--Agentic Tool & 25.04, 28.51 & 78.36, 25.39 & -12.40 & 158 & -1.88 & $< .001$ & [-2.3, -1.3] \\
       & Agentic Tool--LLM & 78.36, 25.39 & 68.69, 30.91 & -2.25 & 158 & -0.34 & .026 & [-0.7, 0.0] \\
\midrule
Blame  & Human--LLM & 28.23, 32.88 & 69.94, 31.45 & -9.32 & 158 & -1.37 & $< .001$ & [-1.8, -0.9] \\
       & Human--Agentic Tool & 28.23, 32.88 & 79.25, 26.23 & -11.41 & 158 & -1.68 & $< .001$ & [-2.1, -1.2] \\
       & Agentic Tool--LLM & 79.25, 26.23 & 69.94, 31.45 & -2.08 & 158 & -0.30 & .039 & [-0.6, 0.0] \\
\midrule
Foreseeability & Human--LLM & 17.91, 25.91 & 57.31, 30.95 & -10.60 & 158 & -1.35 & $< .001$ & [-1.7, -0.9] \\
               & Human--Agentic Tool & 17.91, 25.91 & 63.54, 30.33 & -12.28 & 158 & -1.56 & $< .001$ & [-1.9, -1.1] \\
               & Agentic Tool--LLM & 63.54, 30.33 & 57.31, 30.95 & -1.67 & 158 & -0.21 & .096 & [-0.5, 0.0] \\
\midrule
Counterfactual & Human--LLM & 31.35, 32.12 & 24.05, 29.06 & 2.08 & 158 & 0.23 & .078 & [0.0, 0.5] \\
               & Human--Agentic Tool & 31.35, 32.12 & 20.74, 29.11 & 3.24 & 158 & 0.34 & .009 & [0.0, 0.6] \\
               & Agentic Tool--LLM & 20.74, 29.11 & 24.05, 29.06 & 0.94 & 158 & 0.10 & .347 & [-0.1, 0.3] \\
\bottomrule
\multicolumn{9}{l}{\footnotesize Note. Holm correction applied to $p$-values.}
\end{tabular}
}
\end{table}

\subsubsection{Independent-Samples t-Tests for Each Agent Across Conditions}

To examine the effect of the experimental manipulation between the low and medium agency conditions in Study 5, we conducted pairwise comparisons for each agent across the four measures, as reported in Table~\ref{tabappendixStudy53}. Overall, for causality and blame, all pairwise differences were significant. For foreseeability, the difference was significant for the human and the agentic tool, but not for the LLM component. For counterfactual judgments, only the human showed a significant difference, whereas the LLM and agentic tool components did not.

\begin{table}[h!]
\centering
\caption{Pairwise comparisons between Low and Medium Agency conditions for each agent across measures.}
\label{tabappendixStudy53}
\resizebox{\textwidth}{!}{%
\begin{tabular}{c c c c c c c c c}
\toprule
Agent & Measure & Low ($M, SD$) & Medium ($M, SD$) & $t$ & $df$ & $dz$ & $p$ & 95\% CI \\
\midrule
Human & Cause & 88.73, 14.87 & 25.04, 28.51 & 17.74$^{\dagger}$ & 118.7 & 2.80 & <.001 & [2.3, 3.2] \\
LLM & Cause & 51.83, 33.42 & 68.69, 30.91 & -3.32 & 159 & -0.52 & .001 & [-0.8, -0.2] \\
Agentic tool & Cause & 56.75, 33.08 & 78.36, 25.39 & -4.65$^{\dagger}$ & 149.9 & -0.73 & <.001 & [-1.0, -0.4] \\
\midrule
Human & Blame & 92.86, 13.83 & 28.23, 32.88 & 16.22$^{\dagger}$ & 105.8 & 2.56 & <.001 & [2.0, 3.0] \\
LLM & Blame & 48.06, 33.68 & 69.94, 31.45 & -4.25 & 159 & -0.67 & <.001 & [-0.9, -0.3] \\
Agentic tool & Blame & 49.68, 33.52 & 79.25, 26.23 & -6.23$^{\dagger}$ & 151.1 & -0.98 & <.001 & [-1.3, -0.6] \\
\midrule
Human & Foreseeability & 82.07, 21.09 & 17.91, 25.91 & 17.24 & 159 & 2.71 & <.001 & [2.2, 3.1] \\
LLM & Foreseeability & 50.11, 35.88 & 57.31, 30.95 & -1.36$^{\dagger}$ & 156.2 & -0.21 & .175 & [-0.5, 0.0] \\
Agentic tool & Foreseeability & 51.00, 34.80 & 63.54, 30.33 & -2.43$^{\dagger}$ & 156.6 & -0.38 & .016 & [-0.6, -0.07] \\
\midrule
Human & Counterfactual & 20.17, 28.28 & 31.35, 34.12 & -2.26$^{\dagger}$ & 153 & -0.35 & .025 & [-0.6, -0.04] \\
LLM & Counterfactual & 21.15, 23.46 & 24.05, 29.06 & -0.69 & 159 & -0.11 & .486 & [-0.4, 0.1] \\
Agentic tool & Counterfactual & 18.83, 23.13 & 20.74, 29.11 & -0.46 & 159 & -0.07 & .645 & [-0.3, 0.2] \\
\bottomrule
\multicolumn{9}{l}{\footnotesize{$^{\dagger}$ Welch correction applied.}}
\end{tabular}
}
\end{table}

\subsubsection{Mixed ANOVA on the Effect of the AI system Decomposition on Participants’ Judgments}

To test whether decomposing the AI system into its component parts increased perceived causality, blame, and foreseeability relative to the unified presentation of Experiment 2, we conducted a mixed ANOVA with Agent (Human, AI) as a within-subjects factor and Condition (Unified vs. Decomposed) as a between-subjects factor. The dependent variable was the mean score across the four causal judgment questions. In the decomposed condition, the AI system was represented by averaging participants’ ratings of the two AI components (the LLM and the agentic tool) to allow direct comparison with the unified condition. This analysis was conducted separately for the low and medium AI-agency scenarios.

\textbf{Low AI-Agency Condition.} Across causality, blame, and foreseeability there was a significant main effect of Agent, indicating that participants attributed higher scores to humans than to AI systems. Specifically, the effect of Agent was significant for causality, \( F(1, 133) = 76.03,\, p < .001,\, \eta_p^2 = 0.36 \); for blame, \( F(1, 133) = 150.63,\, p < .001,\, \eta_p^2 = 0.53 \); and for foreseeability, \( F(1, 133) = 68.19,\, p < .001,\, \eta_p^2 = 0.33 \). In contrast, the effect was not significant for counterfactual reasoning (\( p = .798 \)). The main effect of Condition (Unified vs.\ Decomposed) was not significant in any of the models, and there was no significant Agent × Condition interaction. Follow-up \( t \)-tests further confirmed that decomposing the AI system did not significantly alter participants’ evaluations for either agent.

\textbf{Medium AI-Agency Condition.} Results revealed a significant main effect of Agent across all measures: causality (\( F(1, 134) = 100.84,\, p < .001,\, \eta_p^2 = 0.429 \)), blame (\( F(1, 134) = 67.21,\, p < .001,\, \eta_p^2 = 0.334 \)), foreseeability (\( F(1, 134) = 107.38,\, p < .001,\, \eta_p^2 = 0.445 \)), and counterfactual reasoning (\( F(1, 134) = 22.11,\, p < .001,\, \eta_p^2 = 0.163 \)). These effects indicate that participants consistently attributed higher scores to the human compared to the AI system across causality, blame, and foreseeability.

A significant main effect of Condition (Unified vs.\ Decomposed) also emerged for causal (\( F(1, 134) = 29.40,\, p < .001,\, \eta_p^2 = 0.180 \)), blame (\( F(1, 134) = 11.89,\, p < .001,\, \eta_p^2 = 0.082 \)), and foreseeability judgments (\( F(1, 134) = 21.63,\, p < .001,\, \eta_p^2 = 0.139 \)), suggesting that participants’ overall attributions differed depending on how the AI system was presented.

Notably, significant Agent × Condition interactions were observed for causality (\( F(1, 134) = 4.38,\, p = .038,\, \eta_p^2 = 0.032 \)), blame (\( F(1, 134) = 4.87,\, p = .029,\, \eta_p^2 = 0.035 \)), foreseeability (\( F(1, 134) = 4.07,\, p = .046,\, \eta_p^2 = 0.030 \)), and counterfactual reasoning (\( F(1, 134) = 6.56,\, p = .012,\, \eta_p^2 = 0.047 \)). Follow-up \( t \)-tests showed that participants rated the AI system as significantly more causal (\( M = 79.50 \)) and with greater foreseeability (\( M = 68.91 \)) in the Unified condition compared to the Decomposed condition (causal: \( M_U = 79.50 \) vs.\ \( M_D = 68.69,\, p = .027 \); foreseeability: \( M_U = 68.91 \) vs.\ \( M_D = 57.31,\, p = .024 \)). Conversely, the human agent received lower scores in the Decomposed condition across all measures (causality: \( M_D = 25.04 \) vs.\ \( M_U = 50.91,\, p < .001 \); blame: \( M_D = 28.23 \) vs.\ \( M_U = 49.73,\, p < .001 \); foreseeability: \( M_D = 17.91 \) vs.\ \( M_U = 42.36,\, p < .001 \); counterfactuals: \( M_D = 24.05 \) vs.\ \( M_U = 19.05,\, p < .001 \)).

\section{Labeling Methodology for Human Responses}
\label{labelmethod}

To analyze the factors underlying the open-ended explanations provided by participants in the experiments, we developed a coding scheme based on the framework of \citet{franklin2022causal}, which we extended to include factors specific to our experiments, such as proximity. 

\begin{itemize}
    \item Role: the role that the agent has in the interaction context (e.g., delegatee, partner). Different roles require distinct skills and levels of competence, as well as the performance of tasks with varying degrees of importance.
    \item Knowledge: the degree to which an agent has the knowledge to subjectively foresees the outcomes of its actions.
    \item Objective foreseeability: represents how likely an outcome really is, irrespective of an agent’s expectations about what is going to happen.
    \item Capability: refers to an agents’ competencies and skills.
    \item Intent, Desires or Aim: Desires have been treated as conceptually separate from intentions in that 1) intentions involve committing to performing the intended actions, while desires do not, 2) intentions are based on reasoning, while desires are an input to this reasoning, and 3) desires can be directed, while intentions are directed at the intender’s own actions.
    \item Autonomy: when an agent is able to make its own decision, without the control of another agent.
    \item Character: Character refers to an agent’s moral character. Is the extent to which the agent is "responsible" or the opposite - "irresponsible". Character thus capture an aspect of agents that is outside of their capacity to be responsible. 
    \item Proximity: References to how direct or close an event is to the outcome in question, i.e., the event’s position in the causal chain (e.g., “actively or physically caused the hack”).
    \item Violation of a norm: References to actions that violate a moral, social, or legal norm, or that are described as deviant or wrong.
    \item Others: None of the above categories apply.
\end{itemize}

%%%%%%%%%%%%%%%%%%%%%%%%%%%%%%%%%%%%%%%%%%%%%%%%%%%%%%%%%%%%

%%% END INSTRUCTIONS %%%

\end{document}